\documentclass{agujournal2019}
\usepackage{url} %this package should fix any errors with URLs in refs.
\usepackage{lineno}
\usepackage[inline]{trackchanges} %for better track changes. finalnew option will compile document with changes incorporated.
\usepackage{soul}
\usepackage{units,gensymb}
\usepackage{diagbox} % pour la dernière table
\newcommand{\cmark}{\ding{51}}%
\newcommand{\xmark}{\ding{55}}%
\usepackage{xparse}
\NewDocumentCommand\Vect{mgg}{
    \mathbf{#1}%
        \IfNoValueF{#3}{_{#3}}%
        \IfNoValueF{#2}{^{#2}}}

\NewDocumentCommand\Ens{mgg}{
    \mathbb{#1}%
        \IfNoValueF{#3}{_{#3}}%
        \IfNoValueF{#2}{^{#2}}}
        
\NewDocumentCommand\Image{mgg}{#1
        \IfNoValueF{#3}{_{#3}}%
        \IfNoValueF{#2}{^{#2}}}
        
\NewDocumentCommand\Mat{mgg}{ 
    \bm{#1}%
        \IfNoValueF{#3}{_{#3}}%
        \IfNoValueF{#2}{^{#2}}}
        
\NewDocumentCommand\Op{mgg}{
    \mathcal{#1}%
        \IfNoValueF{#3}{_{#3}}%
        \IfNoValueF{#2}{^{#2}}}   

\usepackage{rotating}        
\usepackage{comment}          % for Solution 1
\usepackage{amsmath}          % for Solution 1
\usepackage[overload]{empheq} % for Solution 2
\usepackage{systeme}          % for Solution 3
\usepackage{array}            % for Solution 4
\newcolumntype{C}{>{{}}c<{{}}} 
\draftfalse
\usepackage{pifont}% http://ctan.org/pkg/pifont
\usepackage{lipsum}
\journalname{Journal of Advances in Modeling Earth Systems (JAMES)}

\graphicspath{{figures_new_review/}}

\begin{document}

% Usage de trackchanges.sty:
\addeditor{DB} % définir un éditeur
% \note[editor]{The note}
% \annote[editor]{Text to annotate}{The note}
% \add[editor]{Text to add}
% \remove[editor]{Text to remove}
% \change[editor]{Text to remove}{Text to add}

%% ------------------------------------------------------------------------ %%
%  Title
%
% (A title should be specific, informative, and brief. Use
% abbreviations only if they are defined in the abstract. Titles that
% start with general keywords then specific terms are optimized in
% searches)
%
%% ------------------------------------------------------------------------ %%

% Example: \title{This is a test title}

%\title{Supervision methods for deep neural networks interpolation of Sea Surface Height by fusion multi sensors observations. }
%\title{Deep supervised and unsupervised interpolation of Sea Surface Height fields from multi-variate observations }
%\title{Exploring supervision impact on deep interpolation of Sea Surface Height fields from multi-variate observations}
\title{Learning Sea Surface Height Interpolation from Multi-variate Simulated Satellite Observations.}

%\title{Unsupervised Learning of Sea Surface Height Interpolation from Multi-variate Simulated Observations.}
%Exploring Examining supervision impact on deep interpolation of Sea Surface Height fields from multi-variate observations }

%Supervised and unsupervised deep neural networks for interpolation of Sea Surface Height by fusion of multivariate physical observations.}

%% ------------------------------------------------------------------------ %%
%
%  AUTHORS AND AFFILIATIONS
%
%% ------------------------------------------------------------------------ %%

% Authors are individuals who have significantly contributed to the
% research and preparation of the article. Group authors are allowed, if
% each author in the group is separately identified in an appendix.)

% List authors by first name or initial followed by last name and
% separated by commas. Use \affil{} to number affiliations, and
% \thanks{} for author notes.
% Additional author notes should be indicated with \thanks{} (for
% example, for current addresses).

% Example: \authors{A. B. Author\affil{1}\thanks{Current address, Antartica}, B. C. Author\affil{2,3}, and D. E.
% Author\affil{3,4}\thanks{Also funded by Monsanto.}}

\authors{Théo Archambault$^{1,2}$, Arthur Filoche$^{3}$, Anastase Charantonis$^{2,4,5}$, Dominique Béréziat$^{1}$, Sylvie Thiria$^{2}$}
\affiliation{1}{Sorbonne Université, CNRS, LIP6, Paris, France}
\affiliation{2}{Sorbonne Université, CNRS, IRD, MNHN, LOCEAN, Paris, France}
\affiliation{3}{University of Western Australia, Perth, Australia}
\affiliation{4}{ENSIIE, CNRS, LaMME, Evry, France}
\affiliation{5}{Inria, Paris, France}

% \affiliation{2}{Second Affiliation}
% \affiliation{3}{Third Affiliation}
% \affiliation{4}{Fourth Affiliation}

%(repeat as many times as is necessary)

%% Corresponding Author:
% Corresponding author mailing address and e-mail address:

% (include name and email addresses of the corresponding author.  More
% than one corresponding author is allowed in this LaTeX file and for
% publication; but only one corresponding author is allowed in our
% editorial system.)

% Example: \correspondingauthor{First and Last Name}{email@address.edu}

\correspondingauthor{Théo Archambault}{theo.archambault@lip6.fr}

%% Keypoints, final entry on title page.

%  List up to three key points (at least one is required)
%  Key Points summarize the main points and conclusions of the article
%  Each must be 140 characters or fewer with no special characters or punctuation and must be complete sentences

% Example:
% \begin{keypoints}
% \item	List up to three key points (at least one is required)
% \item	Key Points summarize the main points and conclusions of the article
% \item	Each must be 140 characters or fewer with no special characters or punctuation and must be complete sentences
% \end{keypoints}

\begin{keypoints}
%\item This is the preprint of a paper submitted to the Journal of Advances of Modelling Earth Systems (JAMES) in September 2023
\item We developed a realistic simulation of satellite observations of sea surface height and temperature  
\item We compare deep learning supervised and unsupervised strategies to interpolate the sea surface height
%\item We find temperature enhances sea surface height reconstruction, as well as the estimation of the surface currents and mesoscale eddies
\item We present a hybrid learning strategy using supervised training on our simulated data and unsupervised fine-tuning on real-world data 
\end{keypoints}

%\def\Xh{\Vect{X}^{\rm ssh}}
%\def\Xt{\Vect{X}^{\rm sst}}
%\def\Yh{\Vect{Y}{\rm ssh}}
%\def\Yt{\Vect{Y}{\rm sst}}%\def\Hh{\cal{H}^{\rm sst}}
%\def\Ht{\cal{H}^{\rm sst}}
% $Exemples (a supprimer): \Xh, \Yh,\Hh$

%% ------------------------------------------------------------------------ %%
%
%  ABSTRACT and PLAIN LANGUAGE SUMMARY
%
% A good Abstract will begin with a short description of the problem
% being addressed, briefly describe the new data or analyses, then
% briefly states the main conclusion(s) and how they are supported and
% uncertainties.

% The Plain Language Summary should be written for a broad audience,
% including journalists and the science-interested public, that will not have 
% a background in your field.
%
% A Plain Language Summary is required in GRL, JGR: Planets, JGR: Biogeosciences,
% JGR: Oceans, G-Cubed, Reviews of Geophysics, and JAMES.
% see http://sharingscience.agu.org/creating-plain-language-summary/)
%
%% ------------------------------------------------------------------------ %%

%% \begin{abstract} starts the second page

\begin{abstract}
Satellite-based remote sensing missions have revolutionized our understanding of the Ocean state and dynamics. 
Among them, space-borne altimetry provides valuable Sea Surface Height (SSH) measurements, used to estimate surface geostrophic currents. 
Due to the sensor technology employed, important gaps occur in SSH observations. Complete SSH maps are produced using linear Optimal Interpolations (OI) such as the widely-used Data Unification and Altimeter Combination System (\textsc{duacs}). 
On the other hand, Sea Surface Temperature (SST) products have much higher data coverage and SST is physically linked to geostrophic currents through advection.
We propose a new multi-variate Observing System Simulation Experiment (OSSE) emulating 20 years of SSH and SST satellite observations.
We train an Attention-Based Encoder-Decoder deep learning network (\textsc{abed}) on this data, comparing two settings: one with access to ground truth during training and one without.
On our OSSE, we compare \textsc{abed} reconstructions when trained using either supervised or unsupervised loss functions, with or without SST information. We evaluate the SSH interpolations in terms of eddy detection. 
We also introduce a new way to transfer the learning from simulation to observations: supervised pre-training on our OSSE followed by unsupervised fine-tuning on satellite data.
Based on real SSH observations from the Ocean Data Challenge 2021, we find that this learning strategy, combined with the use of SST, decreases the root mean squared error by 24\% compared to OI.
\end{abstract}

\section*{Plain Language Summary}
The surface of the ocean is observed through various sensors embedded in satellites. 
Specifically, the height of the sea surface is a crucial variable as it can be used to estimate surface currents.
It is currently measured through satellite altimeters, but the data acquisition process leaves gaps in their observations. 
Providing fully gridded maps of the sea surface height is thus an important interpolation problem. 
The widely used interpolated product has some troubles, especially when dealing with small and rapidly evolving eddies.
To enhance its quality, we propose an artificial neural network, a trainable method able to estimate complete sea surface height images. 
The flexibility of these methods allows us to use different satellite information, such as the sea surface temperature, which has a better resolution. 
Usually, neural networks are trained on a dataset upon which they learn the link between input and output data. 
However, in a realistic geoscience scenario, the output is never known for sure. We propose a dataset that simulates this problem and explores methodologies to train these methods. Finally, we design a new way of learning SSH reconstruction using both simulated data and real satellite observation.

%% ------------------------------------------------------------------------ %%
%
%  TEXT
%
%% ------------------------------------------------------------------------ %%
\section{Introduction}
\subsection{Background}

Since the first ocean remote sensing missions in the 1970s, satellite observation has become one of the most determining contributions to understanding ocean state and dynamics~\cite{martin_2014}. 
Through the years, satellites have provided a huge amount of data of various physical natures with wide spatial coverage that complemented in situ datasets.
Among these techniques, satellite altimetry is used to retrieve the Sea Surface Height (SSH) a determining variable of the ocean circulation. 
The SSH spatial gradient can be used to estimate geostrophic circulation, i.e. the currents issued from the equilibrium between the Coriolis force and the pressure force in the surface layer of the Ocean. 
SSH (also called Absolute Dynamical Topography by the altimetry community) is currently measured by nadir-pointing altimeters, meaning that they can only take measurements vertically, along their ground tracks, by calculating the return time of a radar pulse. 
This leads to large gaps in the observed SSH, and providing a gap-free product (L4) is a challenging Spatio-Temporal interpolation problem. One of the most widely used L4 products in oceanography applications is provided by the Data Unification and Altimeter Combination System (\textsc{duacs})~\cite{taburet2019}. It is a linear Optimal Interpolation (OI) of the nadir along-track measurements leveraging a covariance matrix tuned on 25 years of data. However, several studies show that \textsc{duacs} OI misses some of the mesoscales structures and eddies~\cite{amores2018,stegner2021}. Improving the reconstruction of gridded altimetry products remains an open challenge.

To enhance the quality of the SSH reconstruction and sea surface current estimation, using additional physical information such as the Sea Surface Temperature (SST) has been demonstrated to be beneficial~\cite{ciani2020, resac,martin2023,archambault2023visapp,fablet2023}. 
SST motion is linked to ocean circulation~\cite{isernfontanet2006}, and therefore to SSH, as currents transport heat in an advection dynamics. 
SST measurements obtained through passive infrared technology offer a remarkably high spatial resolution, ranging from 1.1 to \unit[4.4]{km}~\cite{emery1989avhrr}, even if intermittent clouds introduce data gaps. On the other hand, microwave sensors provide lower-resolution SST data (\unit[25]{km}) which can be obtained through non-raining clouds. Infrared and microwave data are then combined with in situ measurements, to produce fully gridded SST maps~\cite{donlon2012,chin2017}. Thus, a crucial challenge lies in developing efficient reconstruction methods capable of fusing data derived from different remote sensing techniques, each presenting distinct interpolation challenges. This is essential to unlock the full potential of satellite oceanography products.

\subsection{SSH interpolation with deep neural networks}
In the last decade, deep learning has emerged as one of the leading methods to address image inverse problems. Neural networks demonstrated remarkable flexibility in fusing observations from various sources and modalities, exhibiting their capacity to learn complex relationships given enough training samples~\cite{mccann2017,ongie2020}. 
Prior works proved that it is possible to use SST to enhance SSH reconstruction with a deep-learning network, whether from a downscaling perspective~\cite{nardelli2022,resac} or an interpolation one~\cite{fablet2023,martin2023}.
However, training neural networks usually requires fully gridded ground truth, which is unavailable in a realistic geoscience scenario. To overcome this limitation, it is possible to design a twin experiment of the satellite observing system on a numerical simulation, also called an Observing System Simulation Experiment (OSSE). Neural networks can then be trained on simulated data and applied to satellite observations. The Ocean Data Challenge 2020~\cite{DATAch2020} is a 1-year OSSE providing SSH simulated observations and ground truth, aiming to compare various reconstruction methods. Among them, \citeA{fablet2021} performed a supervised deep learning of the SSH interpolation and extended their study using SST showing increased performance~\cite{fablet2023}. However, if the SSH-only network was successfully applied to real data, adapting its SST-using version is still a challenging problem.
Another way to overcome the lack of ground truth is to employ loss functions allowing the neural network to learn from observations alone. \citeA{archambault2023visapp, martin2023} trained a neural network using only SST and SSH observations showing the potential of unsupervised learning for SSH interpolation. This last option has the advantage of not suffering from the domain gap between the simulation and the real data, but we expect unsupervised interpolations to produce less accurate reconstructions.

\subsection{Contributions}
First, as the previously existing Ocean Data Challenge OSSE provided only one year of data without SST, it presents clear limitations to train neural networks. We propose a new OSSE that includes 20 years of SSH and SST data, with realistic simulated observations of these variables.\\

Second, we compare a fixed neural architecture trained in a supervised and unsupervised way, with or without SST. The SSH interpolation is learned by an Attention-Based Encoder-Decoder (ABED) on our OSSE. Its assessment involves evaluating errors in SSH and geostrophic currents reconstruction. Additionally, a comparison of the eddy structures is conducted, both quantitatively and visually.\\

Third, we propose a hybrid learning strategy consisting of supervised pre-training on our OSSE and unsupervised fine-tuning on real-world observations. Specifically, we compare the same network architecture, trained in the three following manners: supervised on our OSSE and directly applied to observations, trained directly on observations, and the proposed hybrid approach.

This paper is structured as follows. In Section~\ref{sec:data}, after giving a rationale for levraging SST information in the interpolation method, we detail our OSSE. In Section~\ref{sec:method} we present our architecture and the training loss functions. In Section~\ref{sec:results} we evaluate the interpolation on our OSSE, in terms of SSH reconstruction and geostrophic circulation errors. We also perform an eddy detection to demonstrate that SST-using methods retrieve more realistic ocean structures, and we compare ourselves to existing state-of-the-art methods on the Ocean Data Challenge 2020 OSSE. Finally, we compare the learning strategies on real observations. In Section~\ref{sec:conclusion}, we discuss the limitations and perspectives of this work.

\section{Multi-variate data}\label{sec:data}

In the following, we provide a rationale for the SSH and SST relationship, outline the reference data source we utilized (Global Ocean physics Reanalysis~\cite{DATAglorys}), and detail our OSSE's SSH and SST observations. We also present the satellite observations that will be used for training and fine-tuning.
 
\subsection{Physical relationship between SSH and SST}
One of the most important uses of SSH data is to recover oceanic currents through geostrophic approximation. It consists of supposing a static equilibrium between the surface projection of the Coriolis force and the resultant pressure forces. Far from the Equator, where the Coriolis force projection is null, it is a good approximation of the circulation. The surface geostrophic currents can be computed from the SSH $h$ following Equation~\ref{eq:geo}
\begin{comment}
Coriolis force by volume: \(\rho f \Vect{k}\wedge\Vect{w}{}{geo} \)

Pressure force by volume: \(- \nabla \Vect{p} \)
In equilibrium: \begin{align}
    \rho f \Vect{k}\wedge\Vect{w}{}{geo} &= - \nabla \Vect{p}\\
    \Leftrightarrow \Vect{w}{}{geo}&=\frac{1}{f\rho}\Vect{k}\wedge\nabla\Vect{p}\\
    \begin{pmatrix}u_{geo} \\[1ex]v_{geo}\\[1ex]0  \end{pmatrix}&= \frac{1}{f\rho}\begin{pmatrix}0 \\[1ex]0\\[1ex]1  \end{pmatrix}\wedge\rho g\begin{pmatrix}\dfrac{\partial p}{\partial x} \\[2ex]\dfrac{\partial p}{\partial y}\\[2ex] \dfrac{\partial p}{\partial z} \end{pmatrix}\\
\end{align}
\end{comment}
\begin{equation}
\Vect{w}{}{\rm geo}=\begin{pmatrix}u_{\rm geo} \\[1ex]
v_{\rm geo} 
\end{pmatrix}=\begin{pmatrix}
-\dfrac{g}{f}\dfrac{\partial h}{\partial y} \\[3ex]
\dfrac{g}{f}\dfrac{\partial h}{\partial x} 
\end{pmatrix}\label{eq:geo}
\end{equation}
where $u_{\rm geo}$ and $v_{\rm geo}$ are the Eastward and Northward geostrophic currents, $x$ and $y$ the Eastward and Northward coordinates and where $f=2\Omega_r\sin(\phi)$ is the Coriolis factor, $\Omega_r$ being the Earth the rotation period, $\phi$ the latitude and $g$ the gravitational acceleration. 

In a first approximation, the surface temperature $T$ can be considered as a passive tracer transported by surface currents. The evolution of a scalar in a velocity field is described by the linear advection given in Equation~\ref{eq:advection}.
\begin{equation}
\dfrac{\partial T}{\partial t}+\Vect{w}.\Vect{\nabla}T=0\label{eq:advection}\end{equation}
Combining the geostrophic and the advection Equations~(\ref{eq:geo},\ref{eq:advection}), we understand why a time series of SST observations should provide pertinent information for constraining the SSH reconstruction. Several studies pointed out the interest in using SST to reconstruct SSH as \citeA{isernfontanet2006,gonzalez2020}, which established spectral relations between SSH and SST in a Surface Quasi Geostrophic framework. 
However, the physical link between temperature and SSH is more complex, as other phenomena must be considered, such as diffusion, convection, circulation between water depths, atmosphere interactions, and viscosity.
Satellite observations of temperature and sea surface height also suffer from instrumental errors and are, by nature, limited to observing the ocean surface. 
This is why neural network architectures, thanks to their flexibility, seem appropriate to learn the complex underlying link between the data.

\subsection{Observing System Simulation Experiment}
To effectively replicate the relationship between the two variables, we propose an Observing System Simulation Experiment (OSSE), meaning a twin experiment that accurately models the satellite observations of the Ocean. This approach is widely used in the geosciences community as it provides
a way to test reconstruction methods and errors~\cite{gaultier2016, amores2018, stegner2021}. With this mindset, SSH and SST variables of a high-resolution simulation are considered as the ground truth ocean state upon which we simulate satellite measurements. The coherence of the relation between SSH and SST is ensured by the physical model, while with our OSSE we produce enough pairs of ground truth/observation to train a neural network.

In this paper, we denote $\Vect{X}{ssh}$ and $\Vect{X}{sst}$ the ground truth fields of the SSH and SST and $\Vect{Y}{ssh}$ and $\Vect{Y}{sst}$, the simulated observations. Hereafter, we detail the reference dataset of our OSSE and the observation operators of the two variables.

\subsubsection{Base simulation }

We conduct our experiments on the Global Ocean Physics Reanalysis product (GLORYS12)~\cite{DATAglorys}. It provides various physical data such as SSH, SST, and oceanic currents with a spatial resolution of 1/12\degree\ (around \unit[8]{km}). GLORYS12 is based on the NEMO 3.6 model~\cite{madec2017nemo} and assimilates satellite observations (SSH along-track observations and SST full domain observations) through a reduced-order Kalman filter. It is updated annually by the Copernicus European Marine Service, making it impossible to use in near real-time applications. We select a temporal subset of this simulation from 2000/03/20/ to 2019/12/29, for a total of 7194 days.

We select a portion of the Gulf Stream, between 33\degree\ to 43\degree\ North and -65\degree\ to -55\degree\ East. This area is known for its intense circulation, its water mass of very different temperatures, and is far enough from the equator that the geostrophic approximation can be applied. Comparing the surface circulation of the model with its geostrophic approximation, we find that an RMSE of \unitfrac[6.6]{cm}{s} for $u_{\rm geo}$ and $6.1$ \unitfrac[6.1]{cm}{s} for $v_{\rm geo}$. Considering the high intensity and variations of the currents in the Gulf Stream (with 37.1 and 34.3 \unitfrac{cm}{s} of standard deviation for $u$ and $v$ respectively), geostrophy seems to be an adequate estimation. Thus, we expect a significant synergy between SSH and SST which a neural network can learn. For computational reasons, we resample the data to images of size $128\times128$ with a bilinear interpolation, corresponding to a resolution of 0.078\degree\ by pixel (approximately \unit[8.7]{km}). Doing so, the perceptive field of the network covers the entire 10\degree\ by 10\degree\ area.

\subsubsection{SSH simulated observations}
\begin{figure}[h!]
    \centering
    \includegraphics[width=1\textwidth]{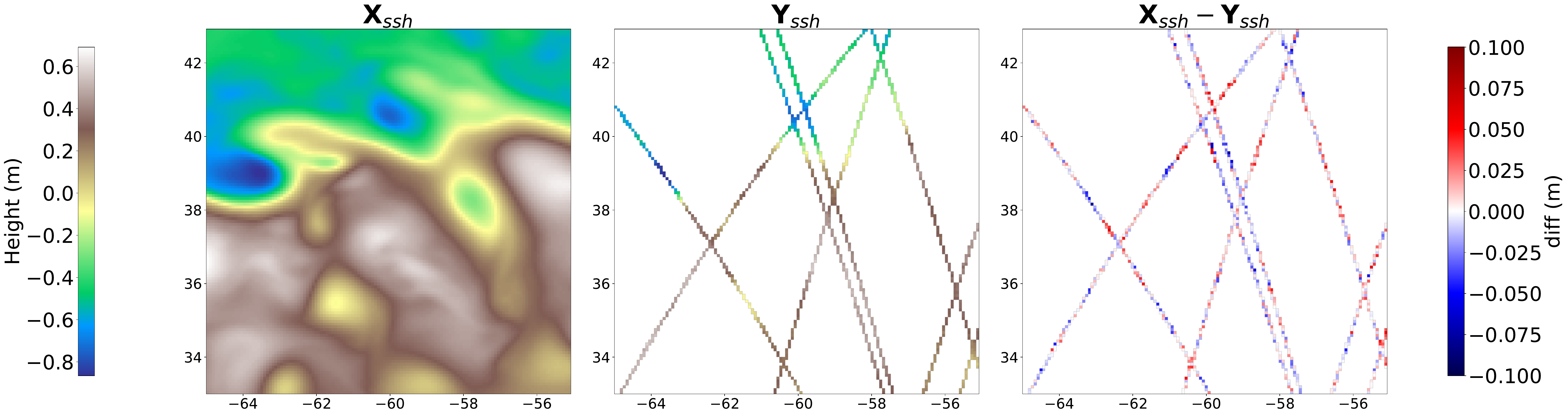}
    \caption{Images of the ground truth SSH from GLORYS12, the simulated along-track measurements, and the difference.}
    \label{fig:data osse ssh}
\end{figure}

The nadir-pointing altimetry satellites take approximately a measurement per second, along their ground tracks. 
Their observations are a series of values with precise spatio-temporal coordinates that we aim to simulate. 
To do so, we retrieve the support of real-world satellite observations denoted $\Vect{\Omega}=\left\{\Omega_i = \left(t_i,lat_i,lon_i\right), i\in \left[0:N\right]\right\}$ from the Copernicus sea level product~\cite{DATAL3tracks}. 
Using $\Vect{\Omega}$ and the ground truth data $\Vect{X}{ssh}$ we simulate SSH observations $\Vect{Y}{ssh}$ as the trilinear interpolation of the simulated field on each point of the support.
We add an instrumental error $\Vect{\varepsilon}\sim\Op{N}\left(0,\sigma\right)$ with $\sigma=\unit[1.9]{cm}$, which is the distribution used in the Ocean data challenge 2020~\cite{DATAch2020}. 
The SSH observations $\Vect{Y}{ssh}$ is defined as following:
\begin{equation}
\Vect{Y}{ssh}=\Op{H}{ssh}\left(\Vect{X}{ssh},\Vect{\Omega}\right)+\Vect{\varepsilon}\label{eq hssh}
\end{equation}

where $\Op{H}{ssh}$ is the trilinear interpolation operator of the ground truth $\Vect{X}{ssh}$ on the support $\Vect{\Omega}$. An example of these simulated along-track measurements is presented on the first row of Figure~\ref{fig:data osse ssh}. For the neural network input observations, we regrid these data to a daily $128\times128$ image. We set the pixel value with no simulated satellite observation to zero, and we average the daily measurements of SSH inside each pixel to represent the mean of the daily data from the different satellites (if any). 
As GLORYS12 reanalysis assimilates along-track SSH data, selecting satellite measurements at the same location as the assimilated data might introduce a bias in our observations. To overcome this issue, we desynchronize the real satellite ground tracks from the one we use to produce SSH observations by introducing a time delay ($772$ days) between the real L3 satellite observations and the simulation. It ensures that simulated along-track data is selected randomly, rather than specifically where the model assimilated real-world observations. 

\subsubsection{SST simulated observations}
SST remote sensing is based on direct infrared imaging, leading to wider measurement swaths but making the data sensitive to cloud cover. 
The so-called L3 satellite products have much higher data coverage, but no observation is possible when clouds are present. 
To fill the gaps, the L3 products from several satellites are merged and interpolated to form the fully gridded image, using complementary microwave satellite sensors (which produce lower resolution data but are less sensitive to clouds), and in situ measurements~\cite{donlon2012,chin2017}.
This results in various resolutions in the same product, where high-resolution structures are artificially smoothed when the cloud cover ($C$) is too thick. 

\begin{figure}[h!]
    \centering
    \includegraphics[width=1\textwidth]{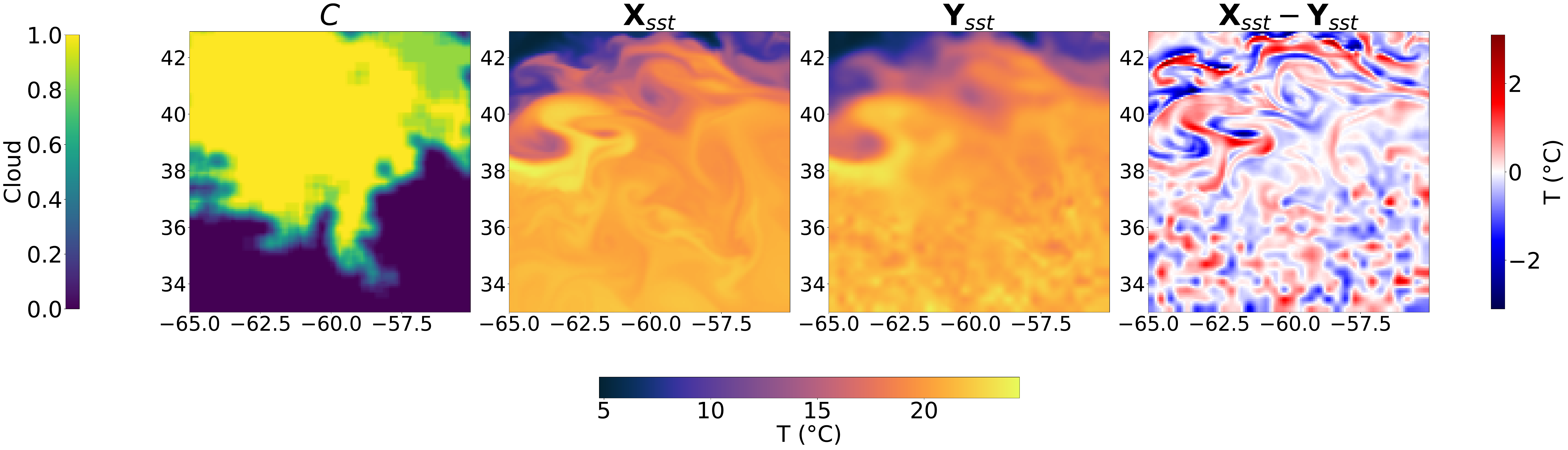}
    \caption{Images of our cloud cover, the ground truth SST from GLORYS12, the noised SST, and the difference.}
    \label{fig: data osse sst}
\end{figure}

We simulate the SST observation operator $\Op{H}{sst}$ as follows:  
\begin{equation}
\Vect{Y}{sst}=\Op{H}{sst}\left(\Vect{X}{sst},C\right)=(1-C)\odot\left(\Vect{X}{sst}+\varepsilon\right)+C\odot{\cal G}_{\sigma_t,\sigma_x}\star\left(\Vect{X}{sst}+\varepsilon\right)\label{eq: hsst}
\end{equation}
where $\odot$ is the element-wise product, $\star$ the convolution product, and $\varepsilon$ is a white Gaussian noise image of size $32\times32$ linearly upsampled to a $128\times128$ image. We also use a spatio-temporal Gaussian filter, $\Op{G}{}{\sigma_t,\sigma_x}$ with $\sigma_t =1.23$ days and $\sigma_x\approx16$(km) to simulate the smoothing of the interpolation performed by satellite products. 
To compute a realistic cloud cover $C$, we use 2 years of data from an NRT L3 product~\cite{DATAcloud}, which we periodically replicate to match the length of our dataset. We then linearly interpolate the cloud cover to our spatial resolution, and perform an average filter with a kernel size approximately equal to $43$ (km). This step is essential, as applying a binary mask results in patches at the frontiers between cloud-free and cloudy regions.
Our SST observations thus present a spatially and temporally correlated noise, with different resolutions depending on cloud cover. In the end, $\Op{H}{sst}$ adds a noise with RMSE of \unit[0.48]{\degree C} where the SST standard deviation of the ground truth is \unit[4.96]{\degree C}, which we present in Figure~\ref{fig: data osse sst}. This observation operator is different from real-world degradations but produces an image with an in-equal noise resolution similar to the errors present in the L4 SST products. Also, as SST presents strong annual variations that should be removed, we deseasonalize it. For each SST image, we subtract the mean image calculated for the corresponding day across the dataset. This is known to improve machine learning time-series prediction~\cite{ahmed2010}, and in our case, it produces better reconstructions as shown in Appendix~\ref{app: deseasonalization}.

%Then a time Gaussian filter (3 days of $\sigma$) is applied to $\varepsilon$ to introduce a temporal correlation.  In order to have 

\subsection{Satellite observations}\label{sec:rwdata}
\begin{figure}[h!]
    \centering
    \includegraphics[width=0.6\textwidth]{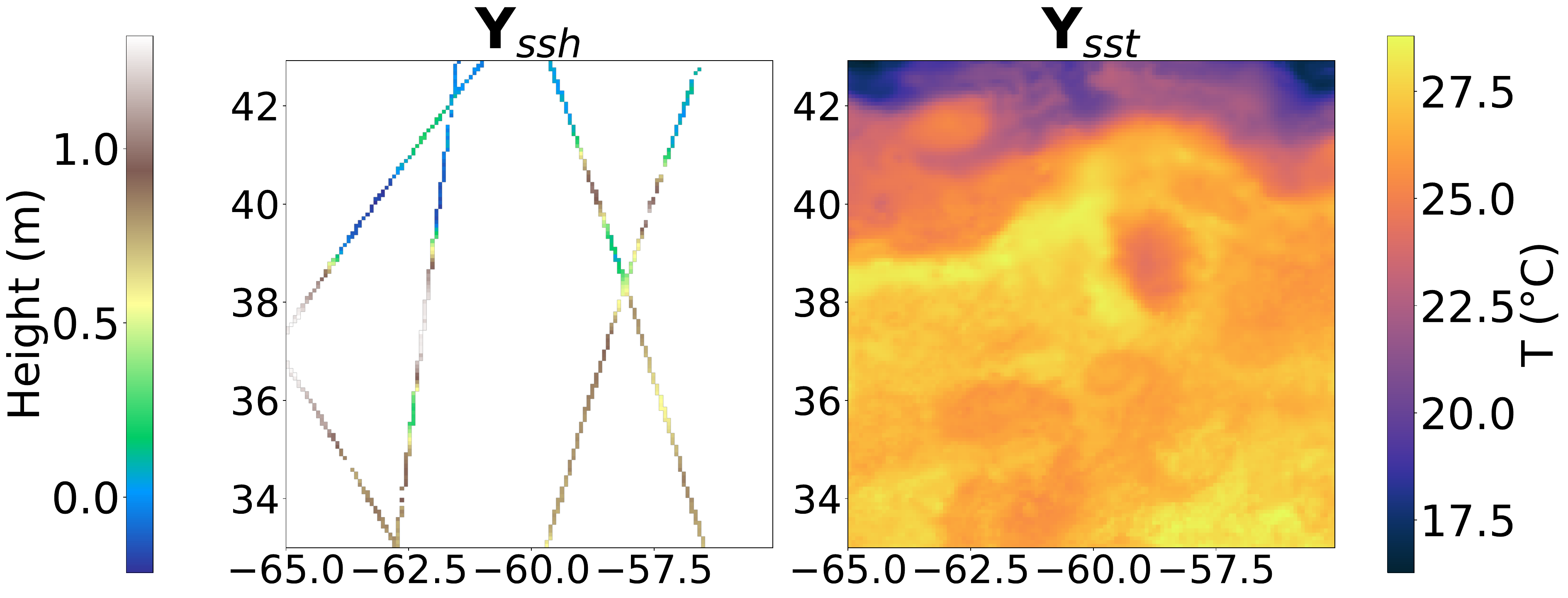}
    \caption{Images of satellite observations of the SSH and the SST, respectively. }
    \label{fig:ose data ssh sst}
\end{figure}

To constitute a dataset of real-world observations, we propose the L3 SSH product that we used to recover realistic satellite ground tracks~\cite{DATAL3tracks}. These data are the inputs used in the \textsc{duacs} optimal interpolation process and are available from the years 1993 to 2023. For the L4 SST product, we use the Multiscale Ultrahigh Resolution (MUR) SST~\cite{DATAmur}. MUR SST is produced through an optimal interpolation of infrared, microwave, and in situ measurements~\cite{chin2017}. Its resolution is very high (0.01\degree), so we linearly interpolate the data to our resolution (0.078\degree), and are available from	2002/05/31 to the present. We select satellite observations from 2002/06/01 to 2022/02/09 for a total of 7194 days which is the same number of timesteps that our OSSE. We also select the same geographical area between 33\degree\ to 43\degree\ North and -65\degree\ to -55\degree\ East. The two data are presented in Figure~\ref{fig:ose data ssh sst}. 

\section{Proposed interpolation method}\label{sec:method}
\subsection{Learning the interpolation}
The observation operator $\Op{H}{ssh}$ previously described can be seen as a forward operator that we aim to inverse. In the past years, deep neural networks, especially convolutional neural networks, have proven their ability to solve ill-posed image inverse problems~\cite{mccann2017} and more specifically inpainting problems~\cite{jam2021,qin2021}. A neural network $f_\theta$ is trained on a database to estimate the true state from observations $f_\theta(y)=\hat{x}$. Learning this inversion operator thus requires $(y,x)$ pairs (supervised) or only $y$ (unsupervised)~\cite{ongie2020}. 

We chose to perform the interpolation on a temporal window of 21 days; the input is thus a tensor of 21 images of SSH, with or without SST images, and the output is the 21 corresponding days of SSH only. 
The neural network estimates the true state from observations, $\Vect{\hat{X}}{ssh}=f_\theta\left(\Vect{Y}\right)$, where $\Vect{Y}=\Vect{Y}{ssh}$ for a SSH-only interpolation, and $\Vect{Y}=\left(\Vect{Y}{ssh},\Vect{Y}{sst}\right)$ if the network uses SST. The length of the time window is discussed in Section~\ref{section:SSHresults}, and training losses of the network in Section~\ref{section:loss}.

\subsection{Architecture}

Convolutional neural networks, one of the most used deep learning methods in image tasks, learn convolution operations able to identify features over space and/or time. These networks have been used for multiple tasks in geosciences, from forecasting~\cite{che2022} to interpolation~\cite{manucharyan2020,fablet2021,martin2023,archambault2023visapp}, and from eddy detection \cite{moschos2020} to super-resolution~\cite{nardelli2022,resac}, to name a few. 
Over time, the machine learning community introduced various ways to organize these convolution operations, each one presenting distinct advantages. Residual layers learn small modifications between their input and output, making neural networks easier to train~\cite{he2016}. 
Attention layers ponder their inputs by a factor between zero and one. This allows subsequent layers to focus on important features while neglecting irrelevant ones, which makes it well-suited to extracting information from contextual variables. It is widely used in many computer vision tasks~\cite{guo2021} and can be transposed to geoscience applications such as~\cite{che2022}. An encoder-decoder architecture progressively compresses and decompresses the input data, identifying structures at different resolutions.

\begin{figure}[h!]
    \centering
    \includegraphics[width=1\textwidth]{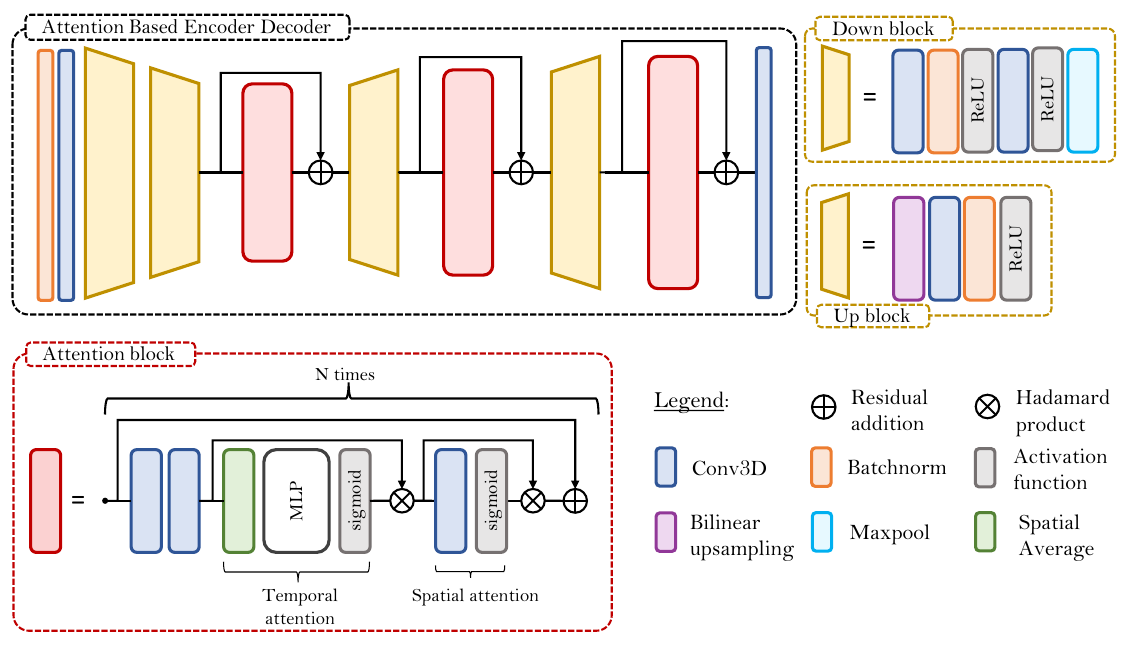}
    \caption{The architecture of the proposed Attention-Based Encoder Decoder (\textsc{abed}) neural network. It is designed to take a time series of 21 images of SSH, with or without a time series of SST. The encoder divides the spatial dimensions of the images by 4 through 2 ``down-block''. Then, a 3D attention layer block is used to highlight relevant information in the images, followed by a residual connection. Finally, a decoding block upsamples the images, and attention and decoding blocks are repeated to get back to the initial image size. }
    \label{fig:network_architecture}
\end{figure}

In this study, we compare different learning techniques on a fixed architecture: an attention-based encoder-decoder (\textsc{abed}) presented in Figure~\ref{fig:network_architecture}. This neural network benefits from the layers described above. The overall structure of our neural network is inspired by~\citeA{che2022}, who introduced a residual U-Net with attention layers for rain nowcasting. We removed U-Net residual connections that were not suited for the interpolation task and changed the attention and the upsampling blocks.
The encoder starts with a batch normalization and a 3D convolution (in time and space) followed by two downsampling blocks that divide spatial dimensions by 2 (see Figure~\ref{fig:network_architecture}). The decoder is composed of residual attention blocks followed by upsampling blocks. 

Hereafter, we describe our attention block, which consists of two essential steps: temporal and spatial attention modules. Our approach builds upon the Convolutional Block Attention Module (CBAM) principle introduced by~\citeA{woo2018}, which successively performs channel and spatial attention. We extend this idea by incorporating temporal information in the channel attention mechanism. To do so, we first compute the spatial average of each channel and instant, resulting in a tensor of size $C\times T$ where $C$ is the channel number and $T$ is the time series length. Subsequently, we apply two one-dimensional convolutional layers with a kernel of size 1, followed by a sigmoid activation function to estimate the attention weights. This corresponds to a 2-layer perceptron shared by every time step, which is different from the CBAM, as it includes the temporal information in the channel attention. These weights are then multiplied to each timestep of every channel, enabling the network to highlight salient features and suppress irrelevant information. After performing temporal attention, we proceed with spatial attention. This step involves utilizing a 3-dimensional convolutional operation, where the kernel size's temporal length matches the time series's length. As a result, the entire time series is aggregated into a single 2D image, which serves as the basis for deriving spatial attention. A residual skip connection is then applied, and the described block is repeated 4, 2, and 1 time for the first, second, and last block, respectively. For further details about our implementation, we provide the PyTorch implementation of our network in \url{https://gitlab.lip6.fr/archambault/james2024}.

\subsection{Loss and regularization}\label{section:loss}
We propose to compare two main strategies to train the neural network. Thanks to the OSSE previously described, we have access to the ground truth, which we can use to learn the interpolation in a classic supervised fashion. However, it is also possible to train directly on observations by applying the $\Op{H}{ssh}$ on the generated map $\Vect{\hat{X}}{ssh}$ before computing the loss (see Equations~\ref{eq :L},\ref{eq: Ltri},\ref{eq : Ltrireg}). \citeA{filoche2022} performed the interpolation with SSH observations only, and, using the same principle, \citeA{archambault2023visapp} showed that it was possible to estimate SSH images starting from SST only and constraining on SSH observations. Both these methods are fitted on one (or a small number) example and must be refitted to be applied to unseen data. Using a larger real-world satellite dataset, \citeA{martin2023} trained a neural network directly from observations by constraining it on independent satellite observations that were not given in the input. However, the lack of ground truth reference makes it harder to compare the different reconstructions, especially regarding detected eddies and structures. We propose to train neural networks using the 3 following losses:
\begin{itemize}
    \item The MSE using ground truth:  
\begin{equation}
\Op{L}{}{\rm sup}(\Vect{X}{ssh},\Vect{\hat{X}}{ssh})= \dfrac{1}{T\times H\times W}\displaystyle\sum_{t,x,y} \left(\Vect{X}{ssh}{t,x,y}-\Vect{\hat{X}}{ssh}{t,x,y}\right)^2\label{eq :L}
\end{equation}

\item The MSE using only observations:  \begin{equation}
\Op{L}{}{\rm unsup}(\Vect{Y}{ssh},\Vect{\hat{X}}{ssh})= \frac{1}{N}\displaystyle\sum_{i} \left(\Vect{Y}{ssh}{i}-\Op{H}{ssh}(\Vect{\hat{X}}{ssh})_{i}\right)^2\label{eq: Ltri}
\end{equation}

\item The MSE using only observations and the regularization introduced by~\citeA{martin2023}:
\end{itemize}
\begin{equation}\label{eq : Ltrireg}
\begin{split}
\Op{L}{}{\rm unsup\_reg}(\Vect{Y}{ssh},\Vect{\hat{X}}{ssh}) =& \Op{L}{}{\rm unsup}(\Vect{Y}{ssh},\Vect{\hat{X}}{ssh})+\lambda_1\frac{1}{N_1}\displaystyle\sum_{i} \left(\frac{\partial }{\partial s}\Vect{Y}{ssh}{i}-\frac{\partial }{\partial s}\Op{H}{ssh}(\Vect{\hat{X}}{ssh})_{i}\right)^2 \\
&+\lambda_2\frac{1}{N_2}\displaystyle\sum_{i} \left(\frac{\partial^2 }{\partial s^2}\Vect{Y}{ssh}{i}-\frac{\partial^2 }{\partial s^2}\Op{H}{ssh}(\Vect{\hat{X}}{ssh})_{i}\right)^2
\end{split}
\end{equation}

where $\frac{\partial }{\partial s}$ is the along-track derivation of the SSH approximated by its rate of change (see Appendix~\ref{app:along track derivative}). $T$ is the temporal length of the time series (here 21), $H$ and $W$ the spatial dimensions of the images (here both equals 128), and $N$, $N_1$, $N_2$, the number of satellite measurements of SSH, SSH first, and SSH second spatial derivative along satellite tracks, respectively. We take  $\lambda_1=\lambda_2=0.05$ the regularization coefficients, the same values used by~\citeA{martin2023}.

\begin{figure}[h!]
    \centering
    \includegraphics[width=0.9\textwidth]{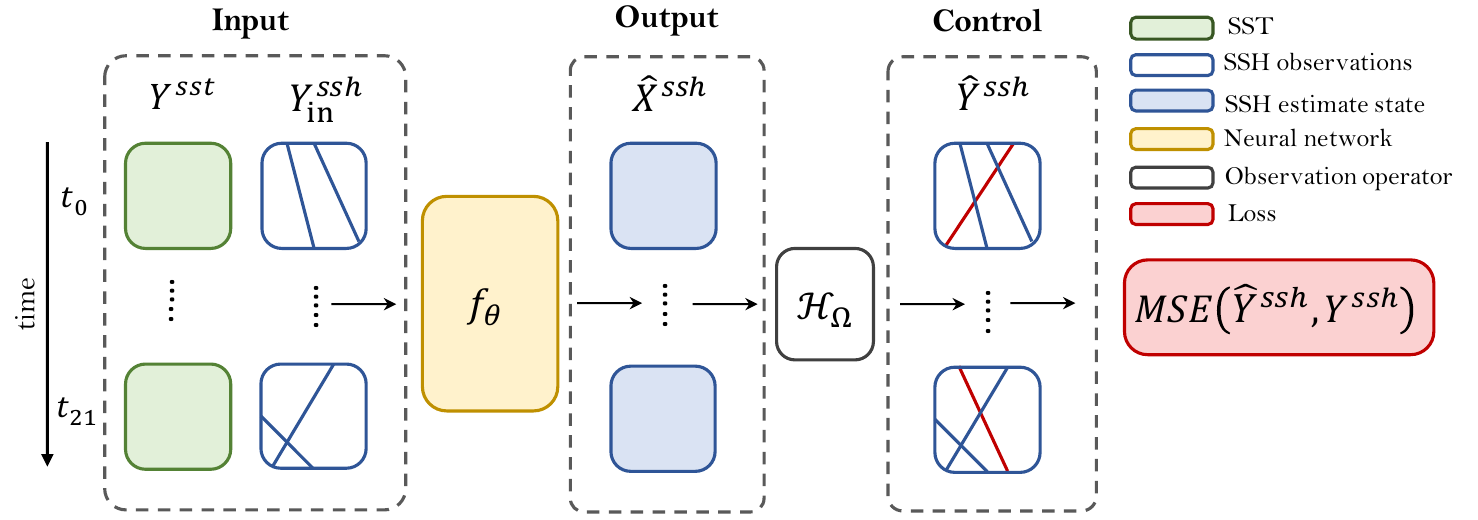}
    \caption{Computational graph of the proposed unsupervised interpolation method. 
    The neural network input is a 21-day time series of SSH satellite observations, excluding data from a single satellite, and optionally includes SST measurements. The network estimates a time series of SSH field states, upon which the observation operator is subsequently applied in order to deduce $\Vect{\hat{Y}}{ssh}$. Finally, the Mean Squared Error between the $\Vect{\hat{Y}}{ssh}$ and $\Vect{Y}{ssh}$ is used to control the network.}
   
    \label{fig: unsupervised inversion computational graph}
\end{figure} 
The losses $\Op{L}{}{\rm unsup}$ and $\Op{L}{}{\rm unsup\_reg}$ apply the observation operator $\Op{H}{ssh}$, before computing the MSE, which allows the training in a framework where only observations are available. Thus, from an interpolation point of view, the inversion methods that use these losses are unsupervised as they can be trained without any ground truth image. However, if we constrain the network on the same observations that were given in input, an over-fitting of along tracks will occur with no guarantee of generalization. 
To avoid this problem, \citeA{martin2023} constrained their network on the observations of one satellite that were withdrawn from the input. Similarly, we remove the data of one satellite from the inputs but we calculate the loss function on all satellite observations (the ones given and the ones left aside). In doing so, the network must generalize outside the along-track measurement that was given as input. In Figure~\ref{fig: unsupervised inversion computational graph}, we call $\Vect{Y}{ssh}{\rm in}$ the input observations and present an unsupervised inversion computational graph.

\subsection{Training details}\label{sec :training details}

\noindent\textbf{Train, validation, test split.} We partitioned the OSSE dataset into three subsets: training, validation, and test data. We used the year 2017 exclusively to test our reconstructions (every analysis conducted in the following was performed on this data). We validate our methods on three distinct time intervals: (1) from 2002/07/14 to 2003/07/28, (2) from 2008/01/05 to 2009/01/18, and (3) from 2013/06/28 to 2014/07/13. The remaining data was used for training, but leaving a 15-day period to prevent data leakage.

\noindent\textbf{Normalization.} We normalize the artificial network's input and output by subtracting the mean and dividing by the standard deviation. The normalization parameters are computed only on the neural network inputs, SST, or along-track data. Specifically, we first perform this normalization for images related to SSH along-track measurements and subsequently replace any missing values with zeros. We normalize the neural network SSH outputs with the statistics computed on the input observations (so that the method remains applicable in an unsupervised setting). When training with the regularized loss of Equation~\ref{eq : Ltrireg}, we also normalize the data from the first and second SSH along-tracks derivative.

\noindent\textbf{Training hyperparameters.} We train every method using an ADAM optimizer~\cite{kingma2017adam} with a learning rate starting at $5.10^{-5}$ and a decay of $0.99$. We perform an early stopping with a patience of 8 epochs. For the supervised training, the stopping criteria is the RMSE of the reconstruction on the fully gridded domain on the validation data, but in the unsupervised setting, we compute this RMSE on left-aside along-track measurements. Doing so, the stopping strategy is still compliant with a situation where no ground truth is accessible.

\noindent\textbf{Ensemble.} As neural network optimization is sensitive to its weight initialization, we train 3 networks for every setting. The so-called ``Ensemble'' estimation is the average SSH map of the 3 networks. An ensemble estimation helps stabilize performances and enhances the reconstruction~\cite{Hinton2015}. In the following, we call ``Ensemble score'' the score of the previously mentioned ensemble estimation and ``Mean score'' the average of the score of each network taken independently. 

\section{Results}\label{sec:results}
In Sections~\ref{section:SSHresults} and \ref{section:eddy}, we compare the different training methods on our OSSE to highlight the drawbacks of unsupervised learning and the advantages of SST. In Section~\ref{section:sota osse}, we assess the similarity of our OSSE and the previously existing one, the Ocean Data Challenge 2020. In Section~\ref{sec:app real_world}, we build upon the conclusions drawn in previous sections to present a transfer learning method from our OSSE to observations.

\subsection{SSH reconstruction and quality of derived geostrophic currents}\label{section:SSHresults}
We compare the fields estimated by the networks trained using the 3 losses $\Op{L}{}{\rm sup}$, $\Op{L}{}{\rm unsup}$ and $\Op{L}{}{\rm unsup\_reg}$, with 3 different sets of input data: only SSH tracks, SSH and the noised SST (denoted nSST), and noise-free SST (denoted SST). The noise-free SST provides an upper-bound performance of the neural network in the case of a perfect physical link between SSH and SST.
We give the RMSE of the SSH estimates fields on the test set in Table~\ref{tab:mercator exp}, and the RMSE on the velocity fields in Table~\ref{tab:mercator exp current}. Systematically, the ensemble reconstruction has a lower RMSE than the mean performance, which is usual in machine learning, as individual member errors are compensated by others. Comparing the ensemble scores, we observe that the supervised loss function outperforms the unsupervised framework in every data scenario. Specifically, in the SSH+SST scenario, the supervised loss decreases the ensemble RMSE of $\Op{L}{}{\rm unsup}$ by 17\%, and 9\% without SST. 
Also, adding SST as an additional input to the network generally improves performance compared to using SSH alone. 
This improvement is observed across all three loss functions, as the error values decrease for SSH+nSST compared to SSH. For instance, the SSH-only ensemble RMSE is decreased by 31\% and 20\% for SST and nSST, respectively, with $\Op{L}{}{\rm sup}$. 
The regularization introduced by~\citeA{martin2023} slightly increases reconstruction but is still close to the unregularized inversion.\\
\begin{table}[h!]
    \centering
    \begin{tabular}{|l|c|c|c|}
        \hline
        Loss & SSH & SSH+nSST & SSH+SST \\
        \hline
        \hline
        $\Op{L}{}{\rm sup}$          & 4.16 | 3.81 & 3.34 | 3.03 & \textbf{2.97} | \textbf{2.63} \\
        $\Op{L}{}{\rm unsup}$        & 4.56 | 4.20 & 3.84 | 3.49 & 3.56 | 3.16  \\
        $\Op{L}{}{\rm unsup\_reg}$   & 4.33 | 4.07 & 3.76 | 3.52 & 3.48 | 3.20  \\
        \hline
    \end{tabular}
    \caption{SSH reconstruction RMSE in centimeters (mean score on the left and ensemble score on the right) of 3 ABED networks. The interpolation is trained using the 3 different losses described in Section~\ref{section:loss} with the following settings: SSH-only interpolation, SSH and noised SST, and SSH and noise-free SST. All metrics are given on the central image of a 21-day time window.  
    }\label{tab:mercator exp}
\end{table}
We estimate the surface currents from the reconstructed SSH from Equation~\ref{eq:geo}, and we compare it to the surface circulation of the model.
The errors on velocity in Table~\ref{tab:mercator exp current} follow the same patterns as the RMSE on the SSH fields but with lesser differences between methods. 
The RMSE is not too far from the minimal error achievable through geostrophy, which is \unitfrac[6.57]{cm}{s} for $u$ and 6.14 for $v$ on this data.

\begin{comment}
  LOSS			 SSH 	 SSH_SST 	 SSH_resSST 	 SSH_bnSST 	 SSH_resbnSST 	
[[4.18 3.19 2.92 3.23 3.23]
 [4.52 3.49 3.62 3.88 3.86]
 [4.38 3.51 3.48 3.7  3.73]]
ENS
[[3.85 2.88 2.59 2.89 2.93]
 [4.16 3.09 3.24 3.5  3.51]
 [4.13 3.26 3.2  3.42 3.48]]
  
\end{comment}

\begin{table}[h!]
    \centering
    \begin{tabular}{|l|c c|c c|c c|}
        \hline
         Loss & \multicolumn{2}{c|}{SSH} &\multicolumn{2}{c|}{SSH+nSST} & \multicolumn{2}{c|}{SSH+SST} \\
        \hline
        \hline
      
          & $u$&$v$ &$u$&$v$ & $u$&$v$ \\
        \hline

        $\Op{L}{}{\rm sup}$        & 12.8 & 13.9 & 11.1 & 12.0 & \textbf{10.1} & \textbf{10.7} \\
        $\Op{L}{}{\rm unsup}$      & 13.4 & 15.5 & 12.0 & 14.1 & 11.1 & 13.1 \\
        $\Op{L}{}{\rm unsup\_reg}$ & 12.8 & 14.3 & 11.7 & 12.9 & 11.0 & 12.0 \\
        \hline

    \end{tabular}
    \caption{Eastward ($u$) and Northward ($v$) surface currents in cm/s. The currents were estimated by applying the geostrophic approximation (see Equation\ref{eq:geo}) on the SSH ensemble estimation of the 3 ABED networks.}
    \label{tab:mercator exp current}
\end{table}

\begin{figure}[h!]
    \centering
    \includegraphics[width=1\textwidth]{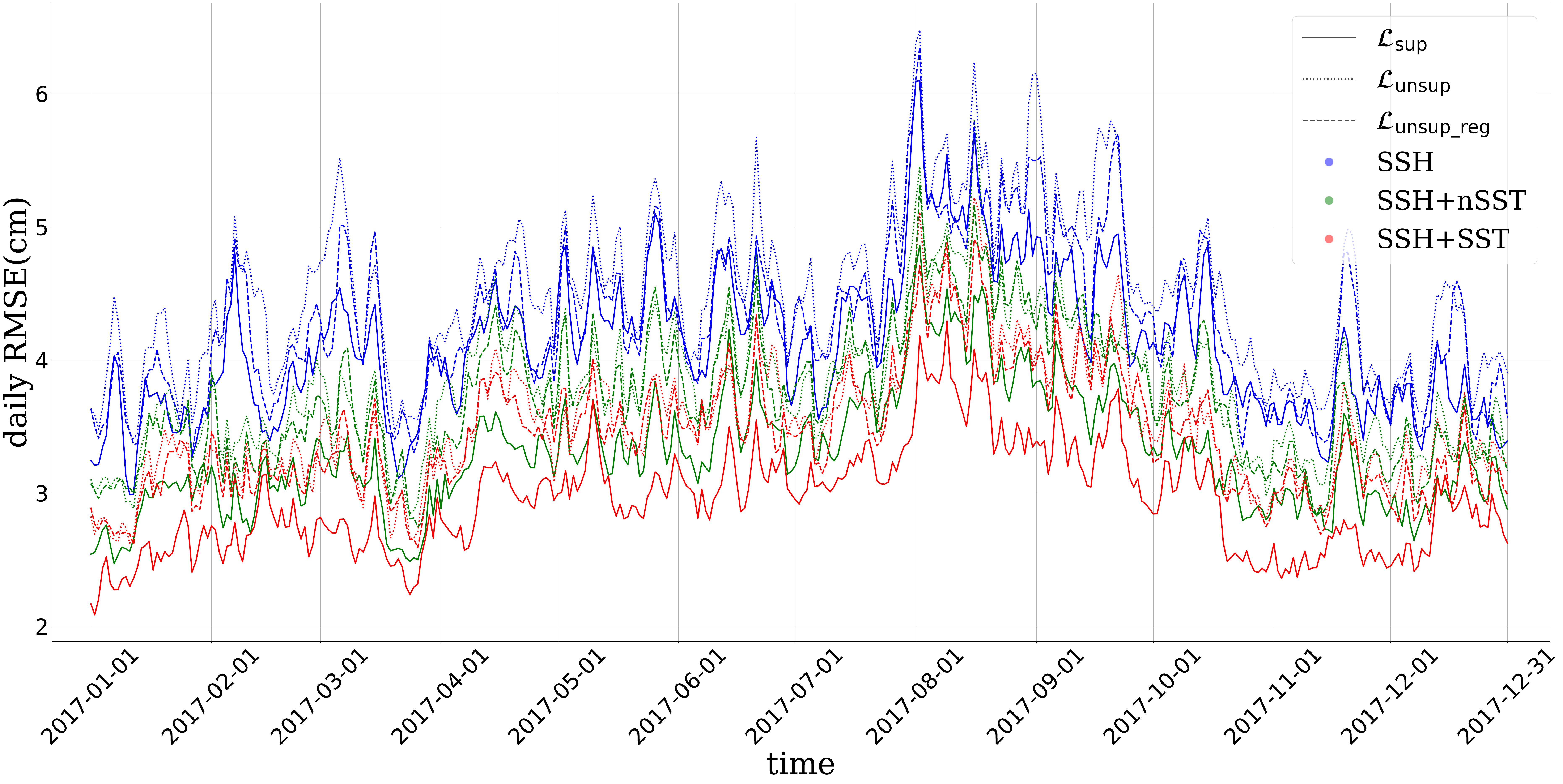}
    \caption{RMSE of the different reconstructions during the test year (2017).}
    \label{fig: time RMSE}
\end{figure}

In Figure~\ref{fig: time RMSE}, we show the daily errors of the different methods on the test year. 
We notice a strong temporal variability of the RMSE, with a notable increase in late summer. 
Specifically, in August and September, all methods are performing worse than in Winter, which can be explained by the high kinetic energy of the ocean in summer~\cite{zhai2008,kang2016}.\\  
\begin{figure}[h!]
    \centering
    \includegraphics[width=1\textwidth]{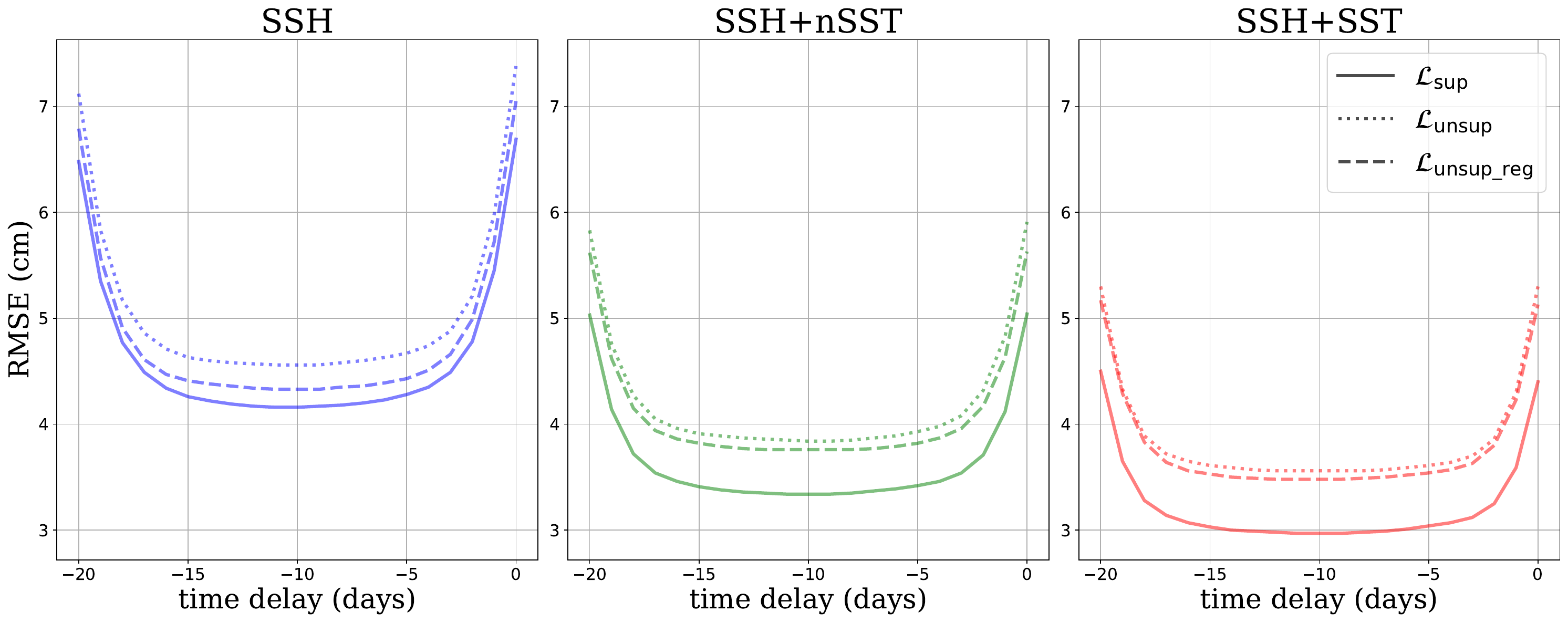}
    \caption{RMSE of the different reconstructions along the time window. The errors at a time delay of $-20$ correspond to an anti-causal scheme (knowing only future observations) whereas $time delay=0$ corresponds to a causal scheme (knowing no future observations). Knowing both past and future observations leads to the optimal reconstruction at $time delay=-10$. }
    \label{fig:window error}
\end{figure} 
An important challenge of ocean satellite products is to provide real-time estimations, as many applications cannot use products available with too much time delay. 
In an operational framework, products that are immediately available are called Near Real Time (NRT) whereas those that require a time delay before release are called Delayed Time (DT). 
While in Table~\ref{tab:mercator exp} we presented the results obtained on the central image of the time window, we can also display their scores along the 21-day temporal window as in Figure~\ref{fig:window error}. 
The central image is a 10-day Delayed Time reconstruction as we need images of observations 10 days in the future.
In Figure~\ref{fig:window error} we can verify that 21 days of data contain enough information to reconstruct the central image: for instance, 5 days from the border of the temporal window the reconstruction error is just 3\% higher than the one at the center. This means that we can significantly reduce the delay (and therefore the training cost of our model) without causing severe drops in performance, which could be useful if applied in an operational framework. However, when it comes to producing NRT products (0 delay) this graph shows that we expect a significant loss of quality in the reconstruction which is usual~\cite{amores2018,stegner2021}.

\subsection{Eddy detection analysis}\label{section:eddy}
\subsubsection{Importance of mesoscale eddies}
Mesoscale eddies play an important role in ocean circulation and dynamics, and their understanding leads to diverse applications in oceanography or navigation~\cite{chelton2011}. Previous studies underline how these structures transport heat, especially between latitudes 0\degree\ and 40\degree\ in the North Atlantic~\cite{jayne2002}, but also salinity~\cite{amores2017}, or plankton~\cite{chelton2011_chl}. In practice, mesoscale eddies and structures are estimated through geostrophic currents derived from satellite altimetry. However, operational satellite products such as \textsc{duacs} OI, have too coarse resolutions to resolve accurately mesoscale structures. Performing an OSSE to simulate the satellite's remote sensing, \citeA{amores2018,stegner2021} showed that \textsc{duacs}-like optimal interpolation aggregates small eddies into larger ones (i.e. with a radius greater than \unit[100]{km}). These interpolations also capture a small percentage of eddies in the model simulation (around 6\% in the North Atlantic) and change the eddies' distribution and properties. This is why we are interested in finding to what extent our reconstruction methods can detect small eddies in the ground truth, and how well the detected eddies are resolved and their physical properties conserved.
 
\subsubsection{Automatic eddy detection algorithm: AMEDA}
We use the Angular Momentum for Eddy Detection and tracking Algorithm (AMEDA) introduced by~\citeA{AMEDA} to perform the eddies detection. It is based on the Local Normalized Angular Momentum (LNAM), a dynamic metric first introduced by~\cite{mkhinini2014}, that we define hereafter:
\begin{equation}
\operatorname{LNAM}(P_i)=\frac{\sum_{j}\overrightarrow{P_iP_j}\times\overrightarrow{V_j}}{\sum_{j}\overrightarrow{P_iP_j}.\overrightarrow{V_j}+\sum_{j}|\overrightarrow{P_iP_j}||\overrightarrow{V_j}|}=\frac{L_i}{S_i+BL_i}   \label{eq: LNAM}
\end{equation}
\begin{comment}
$P_i=\big(\begin{smallmatrix} x_i\\y_i\end{smallmatrix}\big)$
$\overrightarrow{P_iP_j}=\big(\begin{smallmatrix} x_j-x_i\\y_j-y_i\end{smallmatrix}\big)$
$\overrightarrow{V_j}=\big(\begin{smallmatrix} Vx_j\\Vy_j\end{smallmatrix}\big)$
$P_i= \begin{pmatrix} x_i\\y_i\end{pmatrix}$
$\overrightarrow{P_iP_j}= \begin{pmatrix} x_j-x_i\\y_j-y_i\end{pmatrix}$ 
$\overrightarrow{V_j}=\begin{pmatrix} Vx_j\\Vy_j\end{pmatrix}$  
\end{comment}

\noindent where $P_i$ is the point of the grid where we compute the $\operatorname{LNAM}$, $P_j$ is a neighbor point of the grid, $\overrightarrow{P_iP_j}$ is the position vector from $P_i$ to $P_j$ and $\overrightarrow{V_j}$ is the velocity vector in $P_j$. Thus, the unnormalized angular momentum $L_i$ is computed through a sum of cross products and is bounded by $BL_i$, so that if $P_i$ is the center of an axisymmetric cyclone (resp anticyclone), $\operatorname{LNAM}(P_i)$ will be equal to 1 (resp -1). Also, if the circulation field is hyperbolic and not an ellipsoid, $S_i$ will reach large values, and $\operatorname{LNAM}(P_i)$ will be close to 0. All sum is computed on a local neighborhood of $P_i$, which is a hyperparameter of the method (typically a square centered in $P_i$). In our case, we used the default parameters where the square has a length of $2\Delta x$, with $\Delta x$ being the grid resolution ($\approx$\unit[9]{km}). 

%A small illustration is presented in Annexe~\ref{app : lnam}.

AMEDA finds potential eddy centers by searching for the local extrema of the $\operatorname{LNAM}$ field, more precisely by taking the points $P_i$ where $|\operatorname{LNAM}(P_i)|>0.7$. The characteristic contour of an eddy is then defined as the closed streamline of maximum velocity which does not include another eddy center. We perform the AMEDA algorithm on the geostrophic velocity field of our estimation and on the ground truth currents. We then look for the eddies that are both present in the ground truth and in our estimation.
An eddy is said to be detected if the distance between its barycenter and the reference one is smaller than the average of the mean radius of the two characteristic contours. This definition allows ``multiple" detection (i.e., colocalization with several eddies). Therefore, we exclude eddies that include more than one candidate in the ground truth. For further details about the AMEDA algorithm, we refer the reader to \citeA{AMEDA}.

\subsubsection{Eddy detection performances}
We present the detection scores of the different reconstruction methods, with three data scenarios and three losses. 
We take the ensemble SSH estimation of the neural networks and perform the AMEDA algorithm on the velocity field derived through the geostrophic approximation (see Equation~\ref{eq:geo}). 

In Table~\ref{tab:mercator exp detection} we present the $F_1$ score, the recall, and the precision of the methods.
The recall tells us the proportion of actual positive instances that were correctly identified by the detection (a recall of $1$ means that all ground truth eddies were detected). 
The precision gauges our trust in the detected eddies (a precision of $1$ means that all eddies in the simulation were also present in the ground truth). 
To aggregate recall and precision, we use the $F_1$ score, which is the harmonic mean of recall and precision. A value of $1$ means a perfect detection: all ground truth eddies were detected, and the estimation produced no false positives.

\begin{table}[h!]\footnotesize
    \centering
    \begin{tabular}{|l||c|c|c||c|c|c||c|c|c|}
        \hline
        Loss & \multicolumn{3}{c||}{SSH} & \multicolumn{3}{c||}{SSH+nSST} & \multicolumn{3}{c|}{SSH+SST} \\
        \hline
        &$F_1$&recall&precision&$F_1$&recall&precision&$F_1$&recall&precision\\
        \hline
        $\Op{L}{}{\rm sup}$& 0.699 & 0.607 & 0.825 & 0.735 & 0.657 & 0.833 &\textbf{ 0.762} & \textbf{0.705} & 0.83\\
        $\Op{L}{}{\rm unsup}$ & 0.692 & 0.634 & 0.76 & 0.715 & 0.665 & 0.772 & 0.731 & 0.694 & 0.771\\
        $\Op{L}{}{\rm unsup\_reg}$& 0.684 & 0.581 & 0.831 & 0.704 & 0.608 & 0.835 & 0.713 & 0.622 & \textbf{0.836}\\
        \hline
     
    \end{tabular}
    \caption{Scores of the AMEDA eddy detection performed on the Ensemble estimation of \textsc{abed} interpolation. The considered scores are the precision, the recall, and the $F_1$ score.}
    \label{tab:mercator exp detection}
\end{table}
\noindent\textbf{Data comparison.} As expected, no matter which loss we consider, the noise-free temperature detection method outperforms the two other scenarios with higher $F_1$ scores.
Even the noisy SST provides important information for eddy reconstruction, as the SSH-only method yields lower results than the two other scenarios. 
We also see that for each loss, the precision scores are less impacted by the input data than the recall is. 
This means that the SSH-only scenario does not produce a lot more false detection than the SST methods but misses much more structures.

\noindent\textbf{Loss comparison.} On the other hand, the loss function used to perform the inversion substantially impacts precision and recall. 
The regularization of the unsupervised loss brings the detection precision to the level of the supervised method (even higher for the SSH-only and SSH+SST) but also reduces the recall of all methods compared to their unregularized version. 
In other words, the regularization prevents the neural network from generating false eddies and from retrieving some structures, which leads to lower $F_1$ scores.
\begin{figure}[h!]
    \centering
    \includegraphics[width=0.92\textwidth]{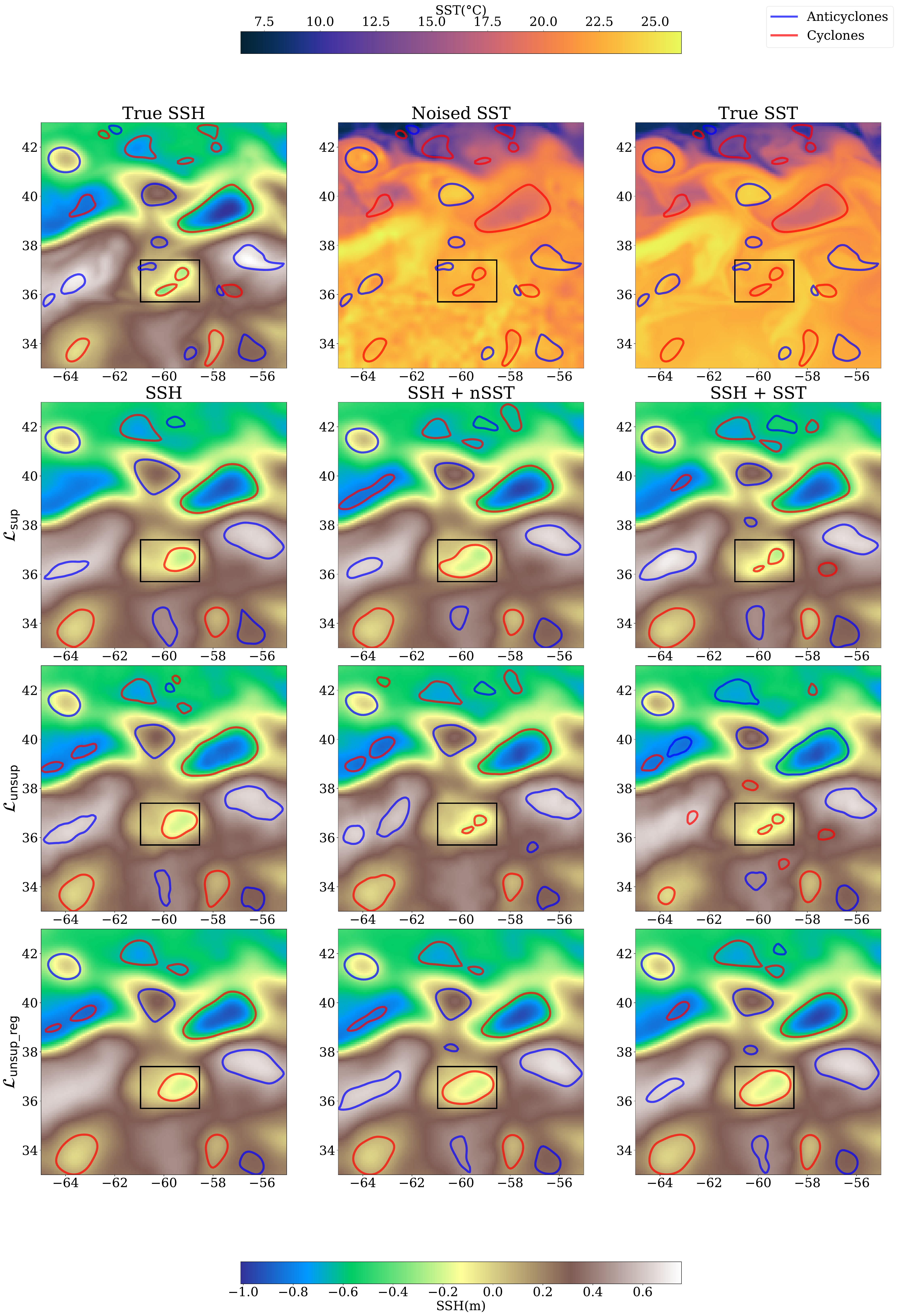}
    \caption{SSH maps and detected eddies the 1$^{st}$ June 2017 on our OSSE. The first line presents the True SSH, the noised SST, and the True SST, on which we plot the eddies detected on the True SSH. The second, third, and last lines present respectively the inversion using $\Op{L}{}{\rm sup}$, $\Op{L}{}{\rm unsup}$, and $\Op{L}{}{\rm unsup\_reg}$. The first, second, and last columns present the maps using the SSH-only, SSH+nSST, SSH+SST data, respectively. Each SSH map is the ensemble reconstruction of 3 networks with their associated eddies.}
    \label{fig: osse plot ssh and eddies}
\end{figure}
\begin{figure}[h!]
    \centering
    \includegraphics[width=0.93\textwidth]{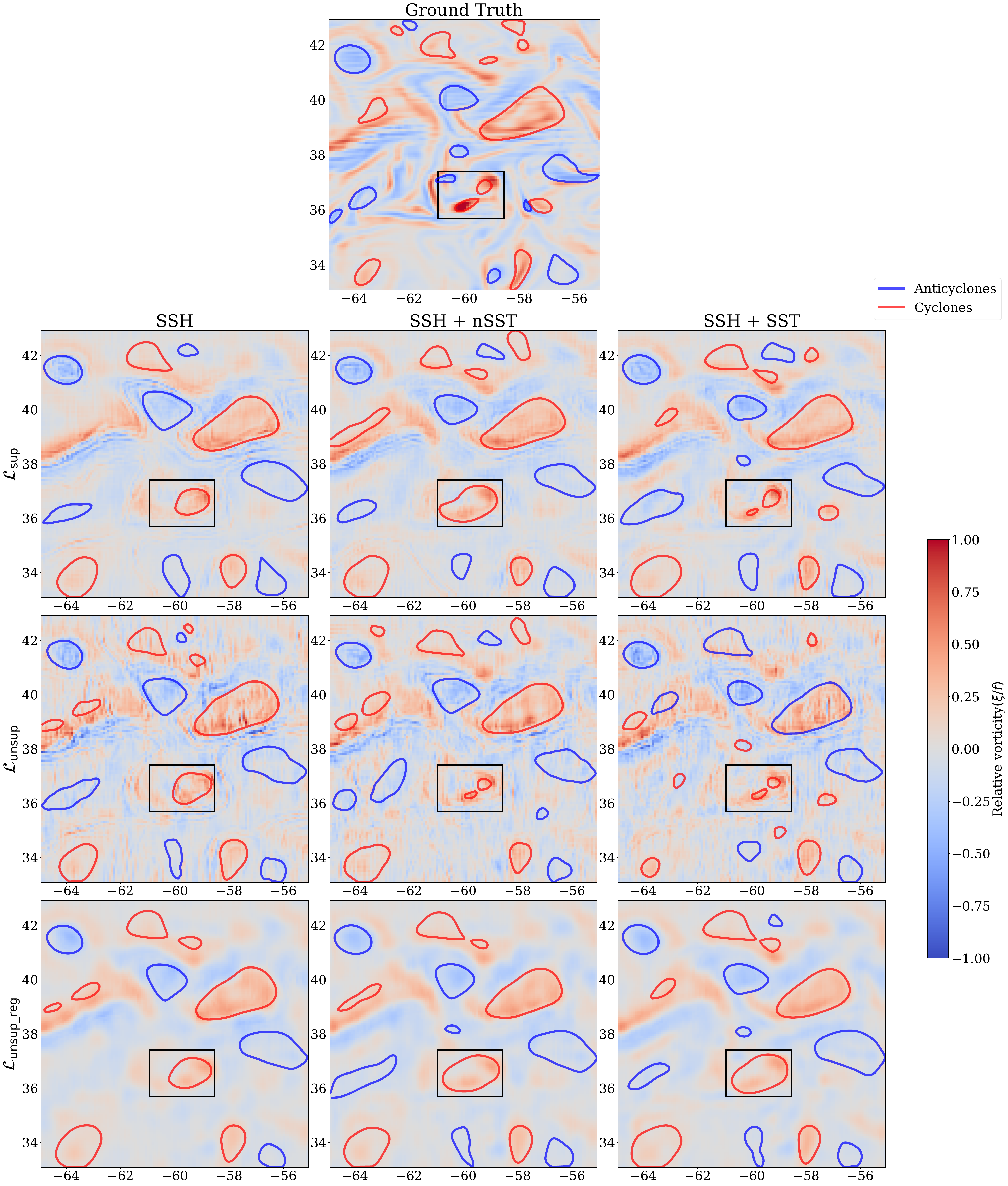}
%\hspace{-1.25cm}\includegraphics[width=0.268\textwidth]{plot_detection_true_vorticity_choice_2017-06-01.pdf}
    %\includegraphics[width=0.93\textwidth]{plot_detection_vorticity_choice_2017-06-01.pdf}
    \caption{Relative vorticity (normalized by the Coriolis factor) and detected eddies on the 1$^{st}$ June 2017 in our OSSE. The first line presents the True relative vorticity. The second, third, and last lines present the neural networks trained with $\Op{L}{}{\rm sup}$, $\Op{L}{}{\rm unsup}$, and $\Op{L}{}{\rm unsup\_reg}$. The first, second, and last columns present the SSH-only, SSH+nSST, and SSH+SST interpolations. Each relative vorticity map is computed from the ensemble SSH estimation of the 3 networks. }
    \label{fig: osse plot vorticity}
\end{figure}

\noindent\textbf{Visual comparison.}
We plot in Figure~\ref{fig: osse plot ssh and eddies} the SSH maps and eddies detected by AMEDA, and in Figure~\ref{fig: osse plot vorticity} the relative vorticity $\xi$ computed from geostrophic currents (see Equation~\ref{eq:geo}) as follows:
\begin{equation}
\label{eq: vorticity}
\xi=\frac{\partial v_{\rm geo}}{\partial x}-\frac{\partial u_{\rm geo}}{\partial y}
\end{equation}
Relative vorticity is an important quantity in the analysis of surface circulation as it highlights areas of important direction change of the stream. $\xi$ is positive in counterclockwise spin and negative in clockwise spin. In the presented figures, we normalize relative vorticity fields by the Coriolis factor $f$.
Figures~\ref{fig: osse plot ssh and eddies},\ref{fig: osse plot vorticity} illustrate an example of the conclusions established in Table~\ref{tab:mercator exp detection}: the SSH-only reconstruction shows fewer eddies than the ones using SST, and aggregates small eddies into larger ones (see highlighted eddies). We also see the effect of regularization, especially in the relative vorticity fields, which are a lot smoother than the ones in the supervised and regularized inversion. This smoothing effect results in a reduced number of detected eddies, as illustrated by the two highlighted eddies that are detected separately when SST is used without regularization.

\subsubsection{Physical properties of detected eddies}

\begin{figure}[h!]
    \centering
    \includegraphics[width=0.92\textwidth]{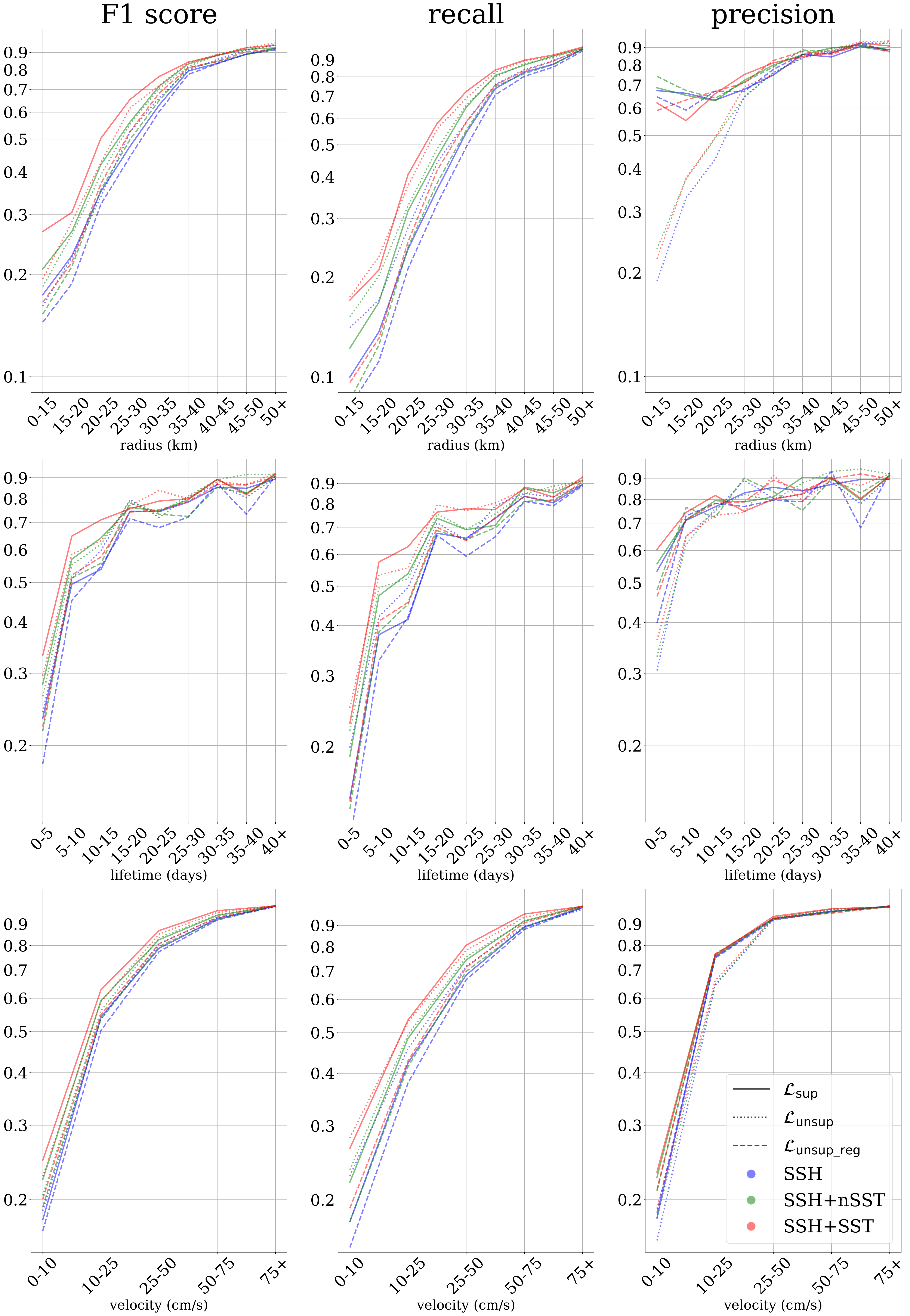}
    \caption{Detection scores of the different methods on eddies separated by radius (first row), lifetime (second row), and maximum velocity (last row). The considered scores are $F_1$ (first column), recall (second column), and precision (third column). The recall tells the proportion of actual positive instances that were correctly identified, the precision gauges the trust that we can put in the detected eddies, and the $F_1$ score aggregates these two values.}
    \label{fig: detection by physical carrac}

\end{figure}

To further investigate the performance of the eddy detection methods, we analyze the detection outcomes based on the physical characteristics of the eddies. 
For instance, smaller eddies tend to have shorter lifespans, making them more challenging to detect due to their decreased likelihood of being observed by satellites. 
Conversely, high-speed eddies are derived from important sea surface height (SSH) variations, thus exhibiting a strong signature in the generated mapping. 
Figure~\ref{fig: detection by physical carrac} shows the detection performances as a function of some key parameters such as maximum radius, lifetime, or maximum velocity along the final closed current line.

As anticipated, using SST and nSST data contributes to the detection of eddies, as indicated by the higher $F_1$ scores achieved in every loss scenario. 
However, small and short-lived eddies are less frequently detected, resulting in lower recall scores. 
Specifically, only 17\% of the eddies with a radius below \unit[15]{km} are successfully detected in the best scenario. Nonetheless, except for the unregularized loss function, the precision scores for the detected eddies remain high, even for small and short-lived ones. 
This observation confirms the previously observed phenomenon where the regularization employed in the inversion process prevents the network from generating false eddy detections but also stops it from capturing a significant portion of the actual eddies. 
This regularization behavior is expected, as forcing a smoothness constraint on the SSH gradient field leads to denying some of the small structures.

We also want to assess the model's accuracy to estimate the eddies' physical properties. 
To this end, we focus on the eddies that were successfully detected by all the methods (3534 eddies out of the 7908 eddies in the ground truth) and compare the physical parameters of the estimated eddies to their values in the corresponding true eddy.
We compute the RMSE and bias of the following parameters: maximum radius and velocity of the characteristic contour of the eddies. 
Once again, Tables~\ref{tab:radius error} and \ref{tab:velocity error} show that SST helps to estimate eddies radius, and velocity. 
Nonetheless, there is a bias of radius and velocity: the size of the eddy is statistically overestimated compared to its ground truth, while its speed is systematically underestimated. 
This is particularly true for the regularized unsupervised loss because of its smoothness constraint, with a velocity bias accountable for half of the RMSE. 

\begin{table}[h!]
    \centering
    \begin{tabular}{|l||c|c||c|c||c|c|}
        \hline
        Loss & \multicolumn{2}{c||}{SSH} & \multicolumn{2}{c||}{SSH+nSST} & \multicolumn{2}{c|}{SSH+SST} \\
        \hline
        &RMSE&bias&RMSE&bias&RMSE&bias\\
        \hline
        $\Op{L}{}{\rm sup}$        & 16.7 & 3.6 & 15.7 & 4.2 & \textbf{14.7 }& 3.7\\
        $\Op{L}{}{\rm unsup}$      & 16.6 & \textbf{0.9} & 16.3 & 1.3 & 15.5 & 1.3\\
        $\Op{L}{}{\rm unsup\_reg}$ & 16.6 & 3.5 & 16.5 & 4.1 & 15.7 & 4.5\\
        \hline
    \end{tabular}
    \caption{Eddies maximum radius RMSE and bias (km). The eddy detection is performed on geostrophic currents of the ensemble estimation and the bias is computed from the estimated radius minus ground truth radius.}
    \label{tab:radius error}
    \begin{tabular}{|l|c|c||c|c||c|c|}
    \hline
    Loss & \multicolumn{2}{c||}{SSH} & \multicolumn{2}{c||}{SSH+nSST} & \multicolumn{2}{c|}{SSH+SST} \\
    \hline
    \hline
    &RMSE&bias&RMSE&bias&RMSE&bias\\
    \hline
    $\Op{L}{}{\rm sup}$        & 14.3 & -5.3 & 12.4 & -3.1 & \textbf{11.7} & \textbf{-2.0}\\
    $\Op{L}{}{\rm unsup}$      & 14.4 & -5.7 & 13.4 & -3.3 & 12.5 & -2.9\\
    $\Op{L}{}{\rm unsup\_reg}$ & 15.2 & -8.3 & 13.9 & -7.4 & 13.2 & -6.5\\
    \hline
    \end{tabular}
    \caption{Eddies maximum velocity RMSE and bias (cm/s).}
    \label{tab:velocity error}
\end{table}

\newpage
\subsection{Comparison with state-of-the-art methods on a NATL60 OSSE}\label{section:sota osse}
Comparing various SSH interpolation methods requires a common benchmark and metrics. The Ocean Data Challenge 2020~\cite{DATAch2020} provides an OSSE similar to the one described in Section~\ref{sec:data}, the state-of-the-art estimations and metrics. The included data are the ground truth SSH, nadir-pointing observations, and a simulation of the SWOT (Surface Water and Ocean Topography) observations, a new altimetry technology~\cite{gaultier2016}. In this study, we have excluded the SWOT measurements as we do not simulate them in our OSSE and focus on nadir-pointing data.
The ground truth is the NATL60 simulation~\cite{ajayi2019spatial} which uses the same physical model (NEMO 3.6)~\cite{madec2017nemo} but at finer scales than GLORYS12, and without assimilation. Given that the NATL60 model also outputs SST and ocean currents fields, we retrieved and used these variables, even though they were not included in the official depository of the challenge.
The state-of-the-art frameworks presented in this challenge are the following:
\begin{itemize}
    \item \textsc{duacs}: the operational linear optimal interpolation leveraging covariance matrix tuned on 25 years of data;
    \item \textsc{dymost}~\cite{ubelmann2016_dymost,ballarotta2020_dymost} and  \textsc{miost}~\cite{ardhuin2020_miost}: two variants of the linear optimal interpolation where the Gaussian covariance model is changed for a non-linear quasi-geostrophic dynamic model (for  \textsc{dymost}) or by a wavelet base (\textsc{miost});
    \item \textsc{bfn}~\cite{guillou2020_bfn}: a data assimilation method that performs a back and forward nudging of a quasi-geostrophic model;
    \item  \textsc{4dv}ar\textsc{n}et~\cite{fablet2021}: a deep learning framework supervised on the Ocean Data Challenge 2020. In this configuration, it only takes SSH observations as input;
    \item  \textsc{musti}~\cite{archambault2023visapp}: an unsupervised neural network fitting SSH along tracks observations starting from an SST image. The fact that this method must be fitted to new observations, limits its operational use.
\end{itemize} 

This benchmark is not complete as the \textsc{c}onv\textsc{ltsm} interpolations introduced by \citeA{martin2023} were trained on real satellite observations only, and the \textsc{4dv}ar\textsc{n}et versions using SST were only computed using SSH observations from nadir pointing satellites and SWOT data~\cite{fablet2023}.
Still, we are interested in evaluating the reconstructions of our networks, trained on our OSSE, on the Ocean Data Challenge 2020 to show the similarity of the two simulated observation systems. To produce our estimation, we regrid the provided data to our resolution (from 0.016° to 0.078°) using trilinear interpolation. We use the SSH simulated observations of the data challenge and the SST of the corresponding NATL60 simulation. The test period includes 42 days of simulation (between 2012/10/22 and 2012/12/02) as defined in the challenge. As such, the comparison is not fully fair since regridding and not training on the same data might bias the scores obtained. It is still a good way to evaluate the similarity of our OSSE to the Ocean Data Challenge 2020, as our approach obtains comparable performances to the state-of-the-art. 
Each method is then evaluated using the following metrics, and we sum up the results in Table~\ref{tab:results natl60}:
\begin{itemize}
    \item $\mu$ and $\sigma_t$ (in cm), are respectively the RMSE of the SSH and the temporal standard deviation of this RMSE. In the data challenge, these two metrics are normalized by the root mean square of the SSH, but we prefer giving it in centimeters to be coherent with the rest of the work;
    \item  $\lambda_x$ (in degrees) and $\lambda_t$ (in days) are two spectral metrics, introduced by~\cite{guillou2020_bfn}. We compute respectively the spatial and temporal power spectrum of the error,  $\lambda_x$ is then the smallest spatial wavelength where the power spectrum of the error is equal to the power spectrum of the signal and  $\lambda_t$ its temporal equivalent. For further information, we refer the reader to~\cite{guillou2020_bfn};
    \item  $\mu_u$ and $\mu_v$ (in cm/s) are the RMSE between the NATL60 currents and the geostrophic currents of the estimation.
\end{itemize}

\begin{table}[h!]
    \centering
    \begin{tabular}{|l|c|c|c|c|c|c|c|c|}
        \hline
         Method & SST&SUP& $\mu$& $\sigma_t$& $\lambda_x$&$\lambda_t$&$\mu_u$&$\mu_v$\\
        \hline
        \hline
        \textsc{duacs}  &\xmark&\xmark &4.89&3.02&1.42&12.08&16.8&16.2\\
        \textsc{dymost}  &\xmark&\xmark&5.18&3.05&1.35&11.87&16.8 &16.8\\
        \textsc{miost}  &\xmark&\xmark&4.21&2.5&1.34&10.34&14.9&14.5\\
        \textsc{bfn}  &\xmark&\xmark&4.7&2.73&1.23&10.64&15.1&15.3\\
        \textsc{4dv}ar\textsc{n}et  &\xmark&\cmark&3.26&1.73&\textbf{0.84}&7.95& 13.1&12.8\\
        \textsc{musti}  &\cmark&\xmark&3.12&1.32&1.23&\textbf{4.14}&12.2&14.2\\
        \hline
        \textsc{abed-ssh} $\Op{L}{}{\rm sup}$               &\xmark&\cmark&3.75&2.0&1.21&8.74&13.3&13.5\\
        \textsc{abed-ssh} $\Op{L}{}{\rm unsup}$          &\xmark&\xmark&4.06&2.19&1.32&9.29&13.7&15.1\\
        \textsc{abed-ssh} $\Op{L}{}{\rm unsup\_reg}$     &\xmark&\xmark&4.23&2.36&1.24&9.98&13.8&14.2\\
        \textsc{abed-ssh-sst} $\Op{L}{}{\rm sup}$           &\cmark&\cmark&\textbf{2.88}&\textbf{1.24}&0.95&4.51&\textbf{11.4}&\textbf{11.4}\\
        \textsc{abed-ssh-sst} $\Op{L}{}{\rm unsup}$      &\cmark&\xmark&3.08&1.41&1.18&5.18&11.8&12.8\\
        \textsc{abed-ssh-sst} $\Op{L}{}{\rm unsup\_reg}$ &\cmark&\xmark&3.39&1.65&1.18&5.7&12.4&12.3\\
        \hline

    \end{tabular}
    \caption{Comparison of the state-of-the-art reconstruction methods on the Ocean Data Challenge 2020. SST stands for whether or not the reconstruction methods are using SST, and SUP stands for whether or not the methods are supervised.   }
    \label{tab:results natl60}
\end{table}
We see in the scores a predominance of neural network-based methods (\textsc{musti}, \textsc{4dv}ar\textsc{n}et and \textsc{abed}) as the importance of the SST in the reconstruction (\textsc{musti}, and \textsc{abed}). The \textsc{abed-ssh} networks do not perform as well as \textsc{4dv}ar\textsc{n}et, but better than optimal interpolations (\textsc{duacs, dymost, miost}) and \textsc{bfn}. %The SST brings substantial improvements to SSH reconstruction in every supervision methodology.
This analysis further supports using SST data in deep learning-based methods for these inverse problems. We can expect around 2 cm of error reduction on the operational interpolation scheme \textsc{duacs} with our best method (41\% of reduction). We also significantly reduce the errors on currents compared to \textsc{duacs}'s, by \unitfrac[5.7]{cm}{s} for $u$ and \unitfrac[5.4]{cm}{s} for $v$ (35\% and 34\% error reduction).

\subsection{Application to real satellite observations}\label{sec:app real_world}

In this section we focus on applying the developed methods to real observations with two objectives in mind: show the utility and realism of our OSSE compared to the pre-existing one, and explore transfer learning strategies. To evaluate our method on a shared benchmark, we use the Ocean Data Challenge 2021~\cite{DATAch2021}, which provides one year of real SSH nadir observations and evaluation metrics. All the evaluations presented in this section are computed on the along-track data from the CryoSat-2 satellite left aside in all the benchmarked methods. The comparison is done on the entire 2017 year, which is the year that we left aside from training on our OSSE to avoid data leakage. To be coherent with the area covered by all the methods, the evaluation area is smaller than the one of the OSSE (between 34\degree\ to 42\degree\ North and -65\degree\ to -55\degree\ East). These real-world measurements present instrumental errors that produce much higher RMSE scores than the ones computed on the OSSE. Also, as we do not have access to complete SSH maps, the metrics used are $\mu$, $\sigma_t$, and $\lambda_x$ (in \unit{km} this time).
For methods requiring SST information, we use satellite SST from~\cite{DATAmur} described in Section~\ref{sec:rwdata}.

\subsubsection{OSSE comparison} 
In this part, we compare the generalization to real satellite data of models trained on our OSSE with models trained on the Ocean Data Challenge 2020.  As this last dataset provides one year of data it can also be used to fit neural networks, but as shown in Appendix~\ref{app:dataset_size_comparison}, training on a longer dataset drastically improves reconstructions. As the existing OSSE does not provide SST data, it is possible to use NATL60 SST, but the lack of realistic noise leads to a domain gap with real data. To this day, if SSH-only neural networks have been successfully transferred to real SSH data, this is not the case for SST-aware ones. %In the following, we will show the advantages of training on our OSSE instead of the Data Challenge 2020. To this end, 
We compare \textsc{abed} trained in a supervised way on our OSSE (SSH-only or using noisy SST), and on the Ocean Data Challenge 2020 (SSH-only or with NATL60 SST output). To train \textsc{abed} on NATL60 data, we regrid the input and target data to our resolution, and use the data split of the challenge~\cite{DATAch2020}; validation of the training between 2012/10/22 and 2012/12/02, and fitting on the remaining days. We use the same hyperparameters as for the training on our OSSE.

Once networks are trained on the simulation, we perform inferences on real data, excluding the tracks from the independent satellite.
\begin{table*}[h!]
    \centering
    \begin{tabular}{|l|c||c|c|c|}
        \hline
         Method &Training OSSE data & $\mu (cm)$& $\sigma_t (cm)$& $\lambda_x (km)$\\
        \hline
        \hline
        \textsc{abed-ssh}&Ocean Data Challenge 2020     &  8.90 | 8.50 & 3.18 | 3.10 & 148 | 143 \\
        \textsc{abed-ssh-sst}&Ocean Data Challenge 2020 & 10.11 | 9.73 & 3.38 | 3.30 & 142 | 137 \\
        \textsc{abed-ssh}& Ours                         &  6.63 | 6.35 & 2.02 | 1.90  & 122 | 119 \\
        \textsc{abed-ssh-sst}& Ours                     &  6.28 | 6.06 & 1.77 | 1.73 & 115 | 113 \\
        \hline

    \end{tabular}
    \caption{Comparison of \textsc{abed} networks trained on our OSSE to the ones trained on the Ocean Data Challenge 2020. All the metrics are computed on independent real data of the Ocean Data Challenge 2021. The left scores are the mean performances on three networks and the right ones are the ensemble scores.   }
    \label{tab: comparison OSSE}
\end{table*}
In Table~\ref{tab: comparison OSSE}, we present the mean and ensemble scores of the models on the Ocean Data Challenge 2021. As expected, \textsc{abed} performs significantly better when trained on our OSSE. Specifically, \textsc{abed-ssh-sst} trained on the Ocean Data Challenge leads to higher errors than its SSH-only version, which shows the domain gap between NATL60 and satellite SST. We conclude that the length of our OSSE and the addition of SST realistic noise enhanced the reconstructions of the real-world SSH.

\subsubsection{Transfer OSSE learning to real-world data.}

Enhancing real-world SSH reconstruction using the information of a simulation is a typical transfer learning problem, where we have access to ground truth in a source domain (OSSE) but not in a target domain (satellite data)~\cite{pan2010}. Given the losses described in Section~\ref{section:loss} and a satellite dataset (see Section~\ref{sec:rwdata}), we can consider three ways to apply our methodology to the Ocean Data Challenge 2021. We partially presented this experiment in~\cite{archambault2024visapp}.

\noindent \textbf{Observation only}: Perform an unsupervised training on real-world data, with the loss function described in Equation~\ref{eq: Ltri}. The training hyperparameters and dataset split are the same as the ones used in the OSSE study (see Section~\ref{sec :training details}).

\noindent \textbf{Simulation only}: Use the networks trained on our OSSE in a supervised way directly on satellite data. As the test year of our OSSE and one of the Ocean Data Challenge 2021 are the same, we have no issues with data leaking. 

\noindent \textbf{Pre-training on OSSE and fine-tuning on satellite data}: After the supervised pre-training on OSSE data we fine-tune the neural network on satellite data for a few epochs using the unsupervised loss. The fine-tuning is done using a small learning rate of $1.10^{-5}$ and a decay of $0.9$. We use an early stopping with a patience of 8 epochs and we save the best model on the validation set. 

%To compare the different methods we use the Ocean Data Challenge 2021~\cite{DATAch2021}, which provides the test and input SSH data and metrics. For SST-using methods, we add the satellite SST~\cite{DATAmur} described in Section~\ref{sec:rwdata}. The metrics of the challenge are computed on the along-track data from the CryoSat-2 satellite for the entire 2017 year.  These real-world measurements present instrumental errors that produce much higher RMSE scores than the ones computed on the OSSE. Also, as we do not have access to complete SSH maps, the metrics used are $\mu$, $\sigma_t$, and $\lambda_x$ (in \unit{km} this time). 

We present in Table~\ref{tab:results sat} the RMSE on the Ocean Data Challenge 2021 of $3$ \textsc{abed} networks trained with the previously mentioned methodologies. One of the first conclusions we can draw from these reconstruction scores is the interest of our OSSE in the training process. The networks fitted on the simulation perform better than their equivalent trained with observations only, except for the network trained using noise-free SST. This shows that our SST noise is realistic, as introducing SST noise during pre-training is beneficial for generalization to satellite data. Secondly, in every data scenario, the pre-trained and fine-tuned networks perform significantly better than their version trained on observation or simulation. In particular, once fine-tuned, the networks pre-trained on nSST and on SST lead to close performance, whereas without fine-tuning, the network trained on noise-free SST produces the worst reconstruction. Given an appropriate fine-tuning strategy, the features learned on noise-free SST that do not apply to satellite data are effectively modified. From this experiment, we conclude that combining supervised training on our OSSE with unsupervised re-fitting on satellite data increases performance, especially if SST is used.
\begin{table}[!h]
    \centering
    \begin{tabular}{|c|c|c|c|}
    \hline
         \diagbox[width=15em]{Learning method}{Input data}
         &  SSH  &SSH+nSST  &SSH+SST \\\hline         
         Observation & 7.07 | 6.75  & 6.63 | 6.27  & | \\
         Simulation &6.63 | 6.35 &  6.28 | 6.06 & 6.89  | 6.68\\
         Pre-training \& Fine-tuning &6.49 | 6.28 & \textbf{6.02} | \textbf{5.82} & 6.04 | 5.84 \\
         \hline
    \end{tabular}
    \caption{Along-track SSH RMSE in centimeters (mean score on the left and ensemble score on the right) of 3 \textsc{abed} networks, computed on 1 year of data provided by the Ocean Data Challenge 2021. The training strategies include observation-only training (with satellite SSH and SSH+nSST), simulation-only training (SSH, SSH+nSST, SSH+SST), and fine-tuned networks (SSH, SSH+nSST, SSH+SST). For the Fine-tuned networks, when a network is pre-trained with noise-free SST, it is still fine-tuned with noisy satellite SST.  }
    \label{tab:results sat}
\end{table} 

In Table~\ref{tab:results sota sat}, we compare our method to the state-of-the-art interpolation methods provided in the context of the Ocean Data Challenge 2021. The included methods are the same as in Table~\ref{tab:results natl60}, plus the \textsc{c}onv\textsc{ltsm-ssh} and \textsc{c}onv\textsc{ltsm-ssh-sst} \cite{martin2023}. We give ensemble scores of the three pre-trained and fine-tuned \textsc{abed} networks using only SSH, or SSH and the noised SST. The enhanced scores of \textsc{abed-ssh-sst} and \textsc{c}onv\textsc{ltsm-ssh-sst} compared to their SSH-only versions emphasize the improvements brought by the SST. \textsc{abed}, \textsc{c}onv\textsc{ltsm} and \textsc{4dv}ar\textsc{n}et lead to better SSH gridding than optimal interpolation-based methods (\textsc{duacs, dymost, miost}) both in terms of RMSE and effective spatial resolution. We also note a significative drop in RMSE score for the \textsc{bfn} method compared to its OSSE reconstruction, which shows that the idealized QG model is less applicable to real-world observations. 
\begin{table*}[h!]
    \centering
    \begin{tabular}{|l|c|c|c|c|c|}
        \hline
         Method & SST&Learning& $\mu (cm)$& $\sigma_t (cm)$& $\lambda_x (km)$\\
        \hline
        \hline
        \textsc{duacs}                      &\xmark&\xmark     & 7.66 & 2.66 & 149 \\
        \textsc{dymost}                     &\xmark&\xmark     & 6.75 & 2.00 & 131 \\
        \textsc{miost}                      &\xmark&\xmark     & 6.75 & 2.00 & 139 \\
        \textsc{bfn}                        &\xmark&\xmark     & 7.46 & 2.59 & 119 \\
        \textsc{4dv}ar\textsc{n}et          &\xmark& Simulation& 6.56 & 1.84 & \textbf{107} \\
        \textsc{musti}                      &\cmark&Observation& 6.26 & 1.96 & 114 \\
        \textsc{c}onv\textsc{ltsm-ssh}      &\xmark&Observation& 6.82 & 1.86 & 114 \\
        \textsc{c}onv\textsc{ltsm-ssh-sst}  &\cmark&Observation& 6.29 & \textbf{1.60} & 108\\
        \hline
        \textsc{abed-ssh}                   &\xmark&Pre-training \& Fine-tuning   &    6.28 & 1.82 & 118 \\
        \textsc{abed-ssh-sst}               &\cmark&Pre-training \& Fine-tuning       & \textbf{5.82} & 1.61 & 108 \\
        \hline

    \end{tabular}
    \caption{Comparison of the state-of-the-art reconstruction methods on the real satellite data of the Ocean Data Challenge 2021. SST stands for whether or not the reconstruction methods are using SST. \textsc{abed-ssh} and \textsc{abed-ssh-sst} stands for the ensemble score of our pre-trained and fine-tuned networks.}
    \label{tab:results sota sat}
\end{table*}

\begin{figure}[h!]
    \centering
    \includegraphics[width=1\textwidth]{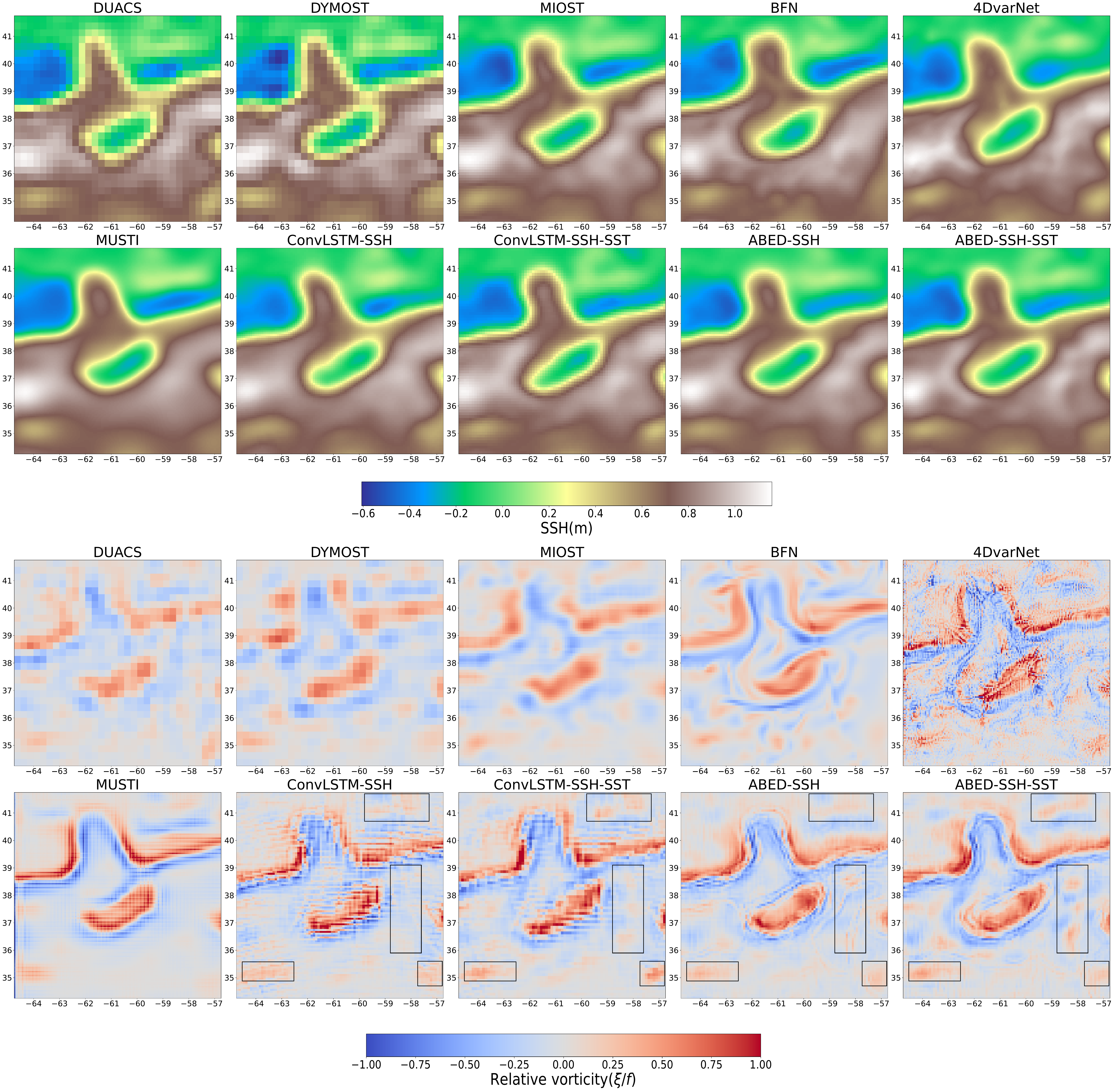}
    \caption{SSH maps and Relative Vorticity maps (normalized by the Coriolis frequency) of the methods from Table~\ref{tab:results sota sat}. The SSH maps are used to compute geostrophic currents from which we derive Relative Vorticity. Due to the different areas covered by the methods, we plot the SSH and RV on a portion of the training area; from 34.25° to 41.75° North and from -64.75° to -56.75° East. On the last four relative vorticity maps, we highlight some regions where small structures are visible in the SST-using interpolations and not visible (or less salient) in their SSH counterparts. }%11 avril 2017
    \label{fig: ose maps}
\end{figure}

In Figure~\ref{fig: ose maps}, we present the SSH maps on the different reconstruction methods with their associated relative vorticity (see Equation~\ref{eq: vorticity}).
The three first methods (\textsc{dymost}, \textsc{duacs}, \textsc{miost}) present smooth vorticity maps as a consequence of the optimal interpolation. All the vorticity maps from neural network-based methods: \textsc{4dv}ar\textsc{n}et, \textsc{musti}, \textsc{c}onv\textsc{ltsm}s, and \textsc{abed}s have higher contrast and also some artifacts due to convolution operations. \textsc{4dv}ar\textsc{n}et in particular, produces very high-frequency variations on which we can see the input satellite path. 
We suppose this is a consequence of the U-Net's skip connections whereas the other networks have Encoder Decoder architectures, less prompt to produce high-frequency noise. For the last four methods, \textsc{c}onv\textsc{ltsm-ssh}, \textsc{c}onv\textsc{ltsm-ssh-sst}, \textsc{abed-ssh}, and \textsc{abed-ssh-sst}, we highlights areas where small structures are visible in the vorticity maps of the SST-using methods but not in their SSH counterparts. The similar shape of the structures between \textsc{c}onv\textsc{ltsm-ssh-sst} and \textsc{abed-ssh-sst} suggests that they are linked to the use of SST and not the deep learning method.

\section{Conclusion and perspectives}\label{sec:conclusion}
\subsection{Summary}
In this work, we designed a new OSSE emulating 20 years of satellite observations of SSH and SST while the previously existing OSSE provided only one year of simulated SSH observations~\cite{DATAch2020}. We were able to train an Attention-Based Encoder Decoder using 3 different loss functions (2 of them learning the reconstruction without ground truth), on three different sets of data (SSH only, SSH and noised SST, SSH, and SST). 
We show a systematic interpolation improvement thanks to the use of SST.
Using temperature data (noisy or not), the unsupervised inversion outperforms even the supervised SSH-only neural network (3\unit[.86]{cm} of RMSE for the unsupervised noisy SST against \unit[4.18]{cm} for the supervised SSH-only method). 
This shows the importance of contextual information to constrain this inverse problem, even while learning with observation only. 

Using AMEDA, an automatic eddy detection algorithm, we were able to identify cyclones and anticyclones in the ground truth and compare them with the eddies detected in the geostrophic approximation of the different mappings. 
This allows a deeper physical interpretation than the SSH reconstruction alone.
We conclude that SST aids in capturing finer structures that might be overlooked by SSH-only methods and that SST-using methods better render the key physical properties of the detected eddies, such as size, speed, or center position. 
Furthermore, in unsupervised reconstruction, we show that the non-regularized and regularized inversions have close detection scores, but their errors are different. 
The regularized inversions exhibited lower recall scores, indicating that certain eddies were not detected due to the smoothing effect of the regularization process. 
However, they demonstrated higher precision scores, implying increased confidence in the successfully detected eddies.

We evaluate \textsc{abed} trained using the data from our OSSE on the Ocean Data Challenge 2020 and compare it with state-of-the-art interpolation techniques. We show that the utilization of SST led to a substantial improvement of 41\% in terms of RMSE for SSH compared to the widely used L4 product from \textsc{duacs}. Moreover, we observed significant improvements of 34\% and 35\% for $u$ and $v$ currents, respectively. These findings present promising perspectives for advancing satellite SSH gridding through the application of deep learning methodologies and the fusion of diverse physical information.

Finally, we presented a novel training strategy, using jointly OSSE and real-world satellite observations. We proposed to perform a transfer from the OSSE to the satellite domain by pre-training the neural network on the OSSE and fine-tuning it on a real-world dataset in an unsupervised way. Comparing the same network trained following three strategies: on simulation only, on observations only, or the one introduced here, we found that using together simulation and satellite data leads to better performances. Specifically, our transfer method achieves state-of-the-art performances on the Ocean Data Challenge 2021, on which we report an RMSE improvement of 24\% compared to \textsc{duacs}.

\subsection{Perspectives}
\textbf{SSH Forecast.}
This study focused on a delayed time interpolation of the SSH. However, near real-time and forecast data are often useful in many operational applications, such as navigation and meteorology. In future works, we would be interested in extending the output window in the future compared to the input one. In doing so, the neural network would be trained to interpolate and forecast the SSH simultaneously. We would be interested to compare a method doing the two tasks simultaneously to a method doing it successively.

\textbf{Global interpolation.}
Furthermore, many challenges still need to be addressed to get toward a global gridded SSH product. For instance, as the geostrophic equilibrium depends on the Coriolis force surface projection and thus on the latitude considered, we may require a model to be trained on several areas with different latitudes. Also, we can wonder which strategy is more efficient between training a global model or several local models, each one specialized for a range of latitude or geographical area. Closed seas and coastal water also have very different physical interactions and might need to be reconstructed by different methods.

\textbf{Using different input and output data.} 
We have demonstrated the benefit of using multi-physical information, specifically SST, to enhance SSH reconstruction through the implementation of a flexible neural network framework. 
The integration of data from diverse physical sources exhibits promising outcomes, yet conventional model-based methods encounter challenges due to noise and observational difficulties associated with real-world data. 
In contrast, machine learning opens doors to augment these methods with diverse and abundant data sources. 
For instance, we employed noisy yet complete SST data in our investigation, but using L3 SST products is also possible. 
Furthermore, an intriguing prospect arises as to whether Level 4 (L4) and Level 3 (L3) SST products can be effectively combined, thereby potentially yielding even more precise and exhaustive information. Other physical measurements might improve the reconstruction, such as chlorophyll maps that track plankton advected by currents~\cite{Kahru2012}.  

\section*{Data availability statement}
The GLORYS12 data \cite{DATAglorys} that we used as a reference throughout this study are freely available and distributed by the European Union-Copernicus Marine Service (\url{https://doi.org/10.48670/moi-00021}). The L3 altimeter~\cite{DATAL3tracks} measurements used to retrieve along tracks coordinates and the L3 SST measurements\cite{DATAcloud} used to compute a realistic cloud cover are distributed by the same service (with doi: \\\url{https://doi.org/10.48670/MOI-00146} and \url{https://doi.org/10.48670/MOI-00164} respectively).

The data of the Ocean Data Challenge 2020 OSSE (ground truth, inputs, and baselines) are available at \url{https://doi.org/10.24400/527896/A01-2020.002} and were developed, validated by CLS and MEOM Team from IGE (CNRS-UGA-IRD-G-INP), France and distributed by Aviso+. To this benchmark data, we add the SST and the surface current of the NATL60 simulation available here:\\\url{https://github.com/CIA-Oceanix/4dvarnet-james-uv-ssc}.

The Satellite dataset is composed of the L3 altimeter measurements previously described~\cite{DATAL3tracks}, and the Multiscale Ultra-High Resolution SST dataset\cite{DATAmur}, available on the following links: \\
\url{https://podaac.jpl.nasa.gov/dataset/MUR25-JPL-L4-GLOB-v04.2}.

The Ocean Data Challenge 2021~\cite{DATAch2021}, data and baselines are available here: \url{https://doi.org/10.24400/527896/a01-2021.005 }%\url{https://github.com/ocean-data-challenges/2021a_SSH_mapping_OSE}
. We add to this benchmark the estimated maps from \cite{martin2023} and \cite{archambault2023visapp}, taken from \url{https://zenodo.org/records/7730739} and\\ \url{https://gitlab.lip6.fr/archambault/visapp2023}, respectively.

The preprocessed data and the weights of our neural networks are available here:\\\url{https://doi.org/10.5281/zenodo.8380280} and our code is hosted on the following repository: \url{https://gitlab.lip6.fr/archambault/james2024}.

\section*{Acknowledgments}
This research was supported by the grant for T.Archambault PhD from Sorbonne Université. The authors acknowledge the AMPHITRITE team, Alexandre Stegner, Briac Le Vu, and Evangelos Moschos, for their useful advice and assistance with the OSSE design and the automatic eddy detection algorithm AMEDA. We sincerely thank the reviewers for their analysis and advice.
\section{Annexes}

\subsection{Along-track spatial derivatives}\label{app:along track derivative}
We calculate the SSH's first and second spatial derivatives along the satellite ground track as described in Equation~\ref{eq: d1} and \ref{eq: d2}. Given $\Vect{Y}{ssh}$, the list of SSH measurements from one satellite (sorted in time), we approximate the derivative by the rate of change of the SSH: 
\begin{equation}
 \frac{\partial }{\partial s}\Vect{Y}{ssh}{i} \simeq \frac{\Vect{Y}{ssh}{i+1}-\Vect{Y}{ssh}{i} }{\triangle s_i} \label{eq: d1}
\end{equation}
\begin{equation}\label{eq: d2}
 \frac{\partial^2 }{\partial s^2}\Vect{Y}{ssh}{i} \simeq \frac{\frac{\partial }{\partial s}\Vect{Y}{ssh}{i+1}-\frac{\partial }{\partial s}\Vect{Y}{ssh}{i} }{\triangle s_i'}
\end{equation}
%
%\noindent 
where $\Vect{Y}{ssh}{i}$ is the $i$-th measurement of SSH, $\triangle s_i$ is the ground distance between the SSH measurements, and $\triangle s_i'$ is the ground distance between the two first derivative approximations.  The lists of first and second spatial derivatives, $\frac{\partial }{\partial s}\Vect{Y}{ssh}{}$ and $\frac{\partial^2 }{\partial s^2}\Vect{Y}{ssh}{}$, are re-centered on new coordinates, corresponding to the dual coordinates of the $\Vect{Y}{ssh}{}$ and  $\frac{\partial }{\partial s}\Vect{Y}{ssh}{}$, respectively.  We only compute the spatial derivatives from observations coming from the same satellite and only if the measurements are taken with less than two seconds of delay. This way we estimate spatial derivatives only where the rate of change is a valid approximation of the derivation. 

\subsection{Impact of the OSSE temporal length on training}\label{app:dataset_size_comparison}
Our OSSE dataset is composed of 7194 days, which leads to 5504 training days once the partition between train, validation, and test sets is made. To evaluate the interest in using more data to constrain the neural network, we train \textsc{abed} network in the optimal configuration (supervised and using noise-free SST). We compare the scenario where all the samples are seen during training with those where only half, a quarter, or a single year of the dataset is used. The validation and test sets remain unchanged, while the training subset is the first consecutive days from the initial training set. Table~\ref{tab: dataset size comparison} presents the RMSE of the reconstructions on the test year of our OSSE. The scores of the networks trained with different dataset sizes clearly show better reconstruction performance when the size increases.  

\begin{table}[h!]
    \centering
    \begin{tabular}{|c|c|c|c|c|}\hline
         Size &Full  & 1/2  & 1/4  &1 year \\\hline\hline
         Number of training samples &5504  & 2752  & 1376  & 365 \\\hline
         RMSE (cm)& 2.97 & 3.97 & 4.77 & 7.76\\\hline
    \end{tabular}
    \caption{Mean RMSE score (in cm) of 3 \textsc{abed} networks trained on our OSSE in a supervised manner using SSH and noise-free SST. We compare the situation where the full, half, a quarter, or one year of the dataset is used. }
    \label{tab: dataset size comparison}
\end{table}

\subsection{Impact of the SST deseasonalization on reconstruction}\label{app: deseasonalization}
In the results presented in this work, we deseasonalized the SST data in the inputs of the neural networks. In Table~\ref{tab: deseasonalization}, we show the RMSE of the neural network using ``native'' SST and the ones using deseasonalized SST. We see that this preprocessing operation decreases the RMSE in every scenario.

\begin{table}[h!]
    \centering
    \begin{tabular}{|l|c|c|}
        \hline
        Loss  & SSH+SST & SSH+SST (deseasonalized) \\
        \hline
        \hline
        $\Op{L}{}{\rm sup}$        & 3.19 | 2.88 & \textbf{2.97} | \textbf{2.63} \\
        $\Op{L}{}{\rm unsup}$      & 3.50 | 3.09 & 3.56 | 3.16  \\
        $\Op{L}{}{\rm unsup\_reg}$ & 3.52 | 3.26 & 3.48 | 3.20  \\
        \hline
    \end{tabular}
    \caption{SSH reconstruction RMSE in centimeters (mean score on the left and ensemble score on the right) of 3 ABED networks. The interpolation is trained using the 3 different losses described in Section~\ref{section:loss} with the following settings:  SSH + noise-free SST and SSH + deseasonalized noised-free SST.
    }
    
    \label{tab: deseasonalization}
\end{table}

\bibliography{biblio}

\begin{thebibliography}{}

\bibitem [\protect \citeauthoryear {%
Ahmed%
, Atiya%
, Gayar%
\BCBL {}\ \BBA {} El-Shishiny%
}{%
Ahmed%
\ \protect \BOthers {.}}{%
{\protect \APACyear {2010}}%
}]{%
ahmed2010}
\APACinsertmetastar {%
ahmed2010}%
\begin{APACrefauthors}%
Ahmed, N\BPBI K.%
, Atiya, A\BPBI F.%
, Gayar, N\BPBI E.%
\BCBL {}\ \BBA {} El-Shishiny, H.%
\end{APACrefauthors}%
\unskip\
\newblock
\APACrefYearMonthDay{2010}{}{}.
\newblock
{\BBOQ}\APACrefatitle {An Empirical Comparison of Machine Learning Models for
  Time Series Forecasting} {An empirical comparison of machine learning models
  for time series forecasting}.{\BBCQ}
\newblock
\APACjournalVolNumPages{Econometric Reviews}{29}{}{594-621}.
\newblock
\begin{APACrefDOI} \doi{10.1080/07474938.2010.481556} \end{APACrefDOI}
\PrintBackRefs{\CurrentBib}

\bibitem [\protect \citeauthoryear {%
Ajayi%
\ \protect \BOthers {.}}{%
Ajayi%
\ \protect \BOthers {.}}{%
{\protect \APACyear {2019}}%
}]{%
ajayi2019spatial}
\APACinsertmetastar {%
ajayi2019spatial}%
\begin{APACrefauthors}%
Ajayi, A.%
, Le~Sommer, J.%
, Chassignet, E.%
, Molines, J\BHBI M.%
, Xu, X.%
, Albert, A.%
\BCBL {}\ \BBA {} Cosme, E.%
\end{APACrefauthors}%
\unskip\
\newblock
\APACrefYearMonthDay{2019}{}{}.
\newblock
{\BBOQ}\APACrefatitle {Spatial and Temporal Variability of North Atlantic Eddy
  Field at Scale less than 100 \unit{km}.} {Spatial and temporal variability of
  north atlantic eddy field at scale less than 100 \unit{km}.}{\BBCQ}
\newblock
\APACjournalVolNumPages{Earth and Space Science Open Archive}{}{}{28}.
\newblock
\begin{APACrefDOI} \doi{10.1002/essoar.10501076.1} \end{APACrefDOI}
\PrintBackRefs{\CurrentBib}

\bibitem [\protect \citeauthoryear {%
Amores%
, Jordà%
, Arsouze%
\BCBL {}\ \BBA {} Le~Sommer%
}{%
Amores%
\ \protect \BOthers {.}}{%
{\protect \APACyear {2018}}%
}]{%
amores2018}
\APACinsertmetastar {%
amores2018}%
\begin{APACrefauthors}%
Amores, A.%
, Jordà, G.%
, Arsouze, T.%
\BCBL {}\ \BBA {} Le~Sommer, J.%
\end{APACrefauthors}%
\unskip\
\newblock
\APACrefYearMonthDay{2018}{{\APACmonth{10}}}{}.
\newblock
{\BBOQ}\APACrefatitle {Up to What Extent Can We Characterize Ocean Eddies Using
  Present-Day Gridded Altimetric Products?} {Up to what extent can we
  characterize ocean eddies using present-day gridded altimetric
  products?}{\BBCQ}
\newblock
\APACjournalVolNumPages{Journal of Geophysical Research:
  Oceans}{123}{}{7220--7236}.
\newblock
\begin{APACrefDOI} \doi{10.1029/2018JC014140} \end{APACrefDOI}
\PrintBackRefs{\CurrentBib}

\bibitem [\protect \citeauthoryear {%
Amores%
, Melnichenko%
\BCBL {}\ \BBA {} Maximenko%
}{%
Amores%
\ \protect \BOthers {.}}{%
{\protect \APACyear {2017}}%
}]{%
amores2017}
\APACinsertmetastar {%
amores2017}%
\begin{APACrefauthors}%
Amores, A.%
, Melnichenko, O.%
\BCBL {}\ \BBA {} Maximenko, N.%
\end{APACrefauthors}%
\unskip\
\newblock
\APACrefYearMonthDay{2017}{{\APACmonth{01}}}{}.
\newblock
{\BBOQ}\APACrefatitle {Coherent mesoscale eddies in the {N}orth {A}tlantic
  subtropical gyre: {3-D} structure and transport with application to the
  salinity maximum} {Coherent mesoscale eddies in the {N}orth {A}tlantic
  subtropical gyre: {3-D} structure and transport with application to the
  salinity maximum}.{\BBCQ}
\newblock
\APACjournalVolNumPages{Journal of Geophysical Research: Oceans}{122}{}{23-41}.
\newblock
\begin{APACrefDOI} \doi{10.1002/2016JC012256} \end{APACrefDOI}
\PrintBackRefs{\CurrentBib}

\bibitem [\protect \citeauthoryear {%
Archambault%
, Filoche%
, Charantonnis%
\BCBL {}\ \BBA {} B{\'e}r{\'e}ziat%
}{%
Archambault%
\ \protect \BOthers {.}}{%
{\protect \APACyear {2023}}%
}]{%
archambault2023visapp}
\APACinsertmetastar {%
archambault2023visapp}%
\begin{APACrefauthors}%
Archambault, T.%
, Filoche, A.%
, Charantonnis, A.%
\BCBL {}\ \BBA {} B{\'e}r{\'e}ziat, D.%
\end{APACrefauthors}%
\unskip\
\newblock
\APACrefYearMonthDay{2023}{{\APACmonth{02}}}{}.
\newblock
{\BBOQ}\APACrefatitle {Multimodal Unsupervised Spatio-Temporal Interpolation of
  satellite ocean altimetry maps} {Multimodal unsupervised spatio-temporal
  interpolation of satellite ocean altimetry maps}.{\BBCQ}
\newblock
\BIn{} \APACrefbtitle {{VISAPP}.} {{VISAPP}.}
\newblock
\APACaddressPublisher{Lisboa, Portugal}{}.
\PrintBackRefs{\CurrentBib}

\bibitem [\protect \citeauthoryear {%
Archambault%
, Filoche%
, Charantonnis%
\BCBL {}\ \BBA {} B{\'e}r{\'e}ziat%
}{%
Archambault%
\ \protect \BOthers {.}}{%
{\protect \APACyear {2024}}%
}]{%
archambault2024visapp}
\APACinsertmetastar {%
archambault2024visapp}%
\begin{APACrefauthors}%
Archambault, T.%
, Filoche, A.%
, Charantonnis, A.%
\BCBL {}\ \BBA {} B{\'e}r{\'e}ziat, D.%
\end{APACrefauthors}%
\unskip\
\newblock
\APACrefYearMonthDay{2024}{{\APACmonth{02}}}{}.
\newblock
{\BBOQ}\APACrefatitle {Pre-training and Fine-tuning Attention Based Encoder
  Decoder Improves Sea Surface Height Multi-variate Inpainting} {Pre-training
  and fine-tuning attention based encoder decoder improves sea surface height
  multi-variate inpainting}.{\BBCQ}
\newblock
\BIn{} \APACrefbtitle {{VISAPP}.} {{VISAPP}.}
\newblock
\APACaddressPublisher{Rome, Italy}{}.
\PrintBackRefs{\CurrentBib}

\bibitem [\protect \citeauthoryear {%
Ardhuin%
\ \protect \BOthers {.}}{%
Ardhuin%
\ \protect \BOthers {.}}{%
{\protect \APACyear {2020}}%
}]{%
ardhuin2020_miost}
\APACinsertmetastar {%
ardhuin2020_miost}%
\begin{APACrefauthors}%
Ardhuin, F.%
, Ubelmann, C.%
, Dibarboure, G.%
, Gaultier, L.%
, Ponte, A.%
, Ballarotta, M.%
\BCBL {}\ \BBA {} Faugère, Y.%
\end{APACrefauthors}%
\unskip\
\newblock
\APACrefYearMonthDay{2020}{{\APACmonth{11}}}{}.
\newblock
\APACrefbtitle {Reconstructing Ocean Surface Current Combining Altimetry and
  Future Spaceborne Doppler Data.} {Reconstructing ocean surface current
  combining altimetry and future spaceborne doppler data.}
\newblock
\APAChowpublished {Earth and Space Science Open Archive}.
\newblock
\begin{APACrefDOI} \doi{10.1002/ESSOAR.10505014.1} \end{APACrefDOI}
\PrintBackRefs{\CurrentBib}

\bibitem [\protect \citeauthoryear {%
Ballarotta%
\ \protect \BOthers {.}}{%
Ballarotta%
\ \protect \BOthers {.}}{%
{\protect \APACyear {2020}}%
}]{%
ballarotta2020_dymost}
\APACinsertmetastar {%
ballarotta2020_dymost}%
\begin{APACrefauthors}%
Ballarotta, M.%
, Ubelmann, C.%
, Rogé, M.%
, Fournier, F.%
, Faugère, Y.%
, Dibarboure, G.%
\BDBL {}Picot, N.%
\end{APACrefauthors}%
\unskip\
\newblock
\APACrefYearMonthDay{2020}{{\APACmonth{09}}}{}.
\newblock
{\BBOQ}\APACrefatitle {Dynamic Mapping of Along-Track Ocean Altimetry:
  Performance from Real Observations} {Dynamic mapping of along-track ocean
  altimetry: Performance from real observations}.{\BBCQ}
\newblock
\APACjournalVolNumPages{Journal of Atmospheric and Oceanic
  Technology}{37}{}{1593-1601}.
\newblock
\begin{APACrefDOI} \doi{10.1175/JTECH-D-20-0030.1} \end{APACrefDOI}
\PrintBackRefs{\CurrentBib}

\bibitem [\protect \citeauthoryear {%
Che%
, Niu%
, Zang%
, Cao%
\BCBL {}\ \BBA {} Chen%
}{%
Che%
\ \protect \BOthers {.}}{%
{\protect \APACyear {2022}}%
}]{%
che2022}
\APACinsertmetastar {%
che2022}%
\begin{APACrefauthors}%
Che, H.%
, Niu, D.%
, Zang, Z.%
, Cao, Y.%
\BCBL {}\ \BBA {} Chen, X.%
\end{APACrefauthors}%
\unskip\
\newblock
\APACrefYearMonthDay{2022}{}{}.
\newblock
{\BBOQ}\APACrefatitle {{ED-DRAP}: Encoder–Decoder Deep Residual Attention
  Prediction Network for Radar Echoes} {{ED-DRAP}: Encoder–decoder deep
  residual attention prediction network for radar echoes}.{\BBCQ}
\newblock
\APACjournalVolNumPages{IEEE Geoscience and Remote Sensing Letters}{19}{}{}.
\newblock
\begin{APACrefDOI} \doi{10.1109/LGRS.2022.3141498} \end{APACrefDOI}
\PrintBackRefs{\CurrentBib}

\bibitem [\protect \citeauthoryear {%
Chelton%
, Gaube%
, Schlax%
, Early%
\BCBL {}\ \BBA {} Samelson%
}{%
Chelton%
, Gaube%
\BCBL {}\ \protect \BOthers {.}}{%
{\protect \APACyear {2011}}%
}]{%
chelton2011_chl}
\APACinsertmetastar {%
chelton2011_chl}%
\begin{APACrefauthors}%
Chelton, D\BPBI B.%
, Gaube, P.%
, Schlax, M\BPBI G.%
, Early, J\BPBI J.%
\BCBL {}\ \BBA {} Samelson, R\BPBI M.%
\end{APACrefauthors}%
\unskip\
\newblock
\APACrefYearMonthDay{2011}{{\APACmonth{10}}}{}.
\newblock
{\BBOQ}\APACrefatitle {The Influence of Nonlinear Mesoscale Eddies on
  Near-Surface Oceanic Chlorophyll} {The influence of nonlinear mesoscale
  eddies on near-surface oceanic chlorophyll}.{\BBCQ}
\newblock
\APACjournalVolNumPages{Science}{334}{}{328--332}.
\newblock
\begin{APACrefDOI} \doi{10.1126/SCIENCE.1208897} \end{APACrefDOI}
\PrintBackRefs{\CurrentBib}

\bibitem [\protect \citeauthoryear {%
Chelton%
, Schlax%
\BCBL {}\ \BBA {} Samelson%
}{%
Chelton%
, Schlax%
\BCBL {}\ \BBA {} Samelson%
}{%
{\protect \APACyear {2011}}%
}]{%
chelton2011}
\APACinsertmetastar {%
chelton2011}%
\begin{APACrefauthors}%
Chelton, D\BPBI B.%
, Schlax, M\BPBI G.%
\BCBL {}\ \BBA {} Samelson, R\BPBI M.%
\end{APACrefauthors}%
\unskip\
\newblock
\APACrefYearMonthDay{2011}{{\APACmonth{10}}}{}.
\newblock
{\BBOQ}\APACrefatitle {Global observations of nonlinear mesoscale eddies}
  {Global observations of nonlinear mesoscale eddies}.{\BBCQ}
\newblock
\APACjournalVolNumPages{Progress in Oceanography}{91}{}{167-216}.
\newblock
\begin{APACrefDOI} \doi{10.1016/J.POCEAN.2011.01.002} \end{APACrefDOI}
\PrintBackRefs{\CurrentBib}

\bibitem [\protect \citeauthoryear {%
Chin%
, Vazquez-Cuervo%
\BCBL {}\ \BBA {} Armstrong%
}{%
Chin%
\ \protect \BOthers {.}}{%
{\protect \APACyear {2017}}%
}]{%
chin2017}
\APACinsertmetastar {%
chin2017}%
\begin{APACrefauthors}%
Chin, T\BPBI M.%
, Vazquez-Cuervo, J.%
\BCBL {}\ \BBA {} Armstrong, E\BPBI M.%
\end{APACrefauthors}%
\unskip\
\newblock
\APACrefYearMonthDay{2017}{}{}.
\newblock
{\BBOQ}\APACrefatitle {A multi-scale high-resolution analysis of global sea
  surface temperature} {A multi-scale high-resolution analysis of global sea
  surface temperature}.{\BBCQ}
\newblock
\APACjournalVolNumPages{Remote Sensing of Environment}{200}{}{154--169}.
\newblock
\begin{APACrefDOI} \doi{https://doi.org/10.1016/j.rse.2017.07.029}
  \end{APACrefDOI}
\PrintBackRefs{\CurrentBib}

\bibitem [\protect \citeauthoryear {%
Ciani%
, Rio%
, Bruno~Nardelli%
, Etienne%
\BCBL {}\ \BBA {} Santoleri%
}{%
Ciani%
\ \protect \BOthers {.}}{%
{\protect \APACyear {2020}}%
}]{%
ciani2020}
\APACinsertmetastar {%
ciani2020}%
\begin{APACrefauthors}%
Ciani, D.%
, Rio, M\BHBI H.%
, Bruno~Nardelli, B.%
, Etienne, H.%
\BCBL {}\ \BBA {} Santoleri, R.%
\end{APACrefauthors}%
\unskip\
\newblock
\APACrefYearMonthDay{2020}{{\APACmonth{05}}}{}.
\newblock
{\BBOQ}\APACrefatitle {Improving the Altimeter-Derived Surface Currents Using
  Sea Surface Temperature ({SST}) Data: A Sensitivity Study to {SST} Products}
  {Improving the altimeter-derived surface currents using sea surface
  temperature ({SST}) data: A sensitivity study to {SST} products}.{\BBCQ}
\newblock
\APACjournalVolNumPages{Remote Sensing}{12}{}{1601}.
\newblock
\begin{APACrefDOI} \doi{10.3390/RS12101601} \end{APACrefDOI}
\PrintBackRefs{\CurrentBib}

\bibitem [\protect \citeauthoryear {%
CLS/MEOM%
}{%
CLS/MEOM%
}{%
{\protect \APACyear {2020}}%
}]{%
DATAch2020}
\APACinsertmetastar {%
DATAch2020}%
\begin{APACrefauthors}%
CLS/MEOM.%
\end{APACrefauthors}%
\unskip\
\newblock
\APACrefYearMonthDay{2020}{}{}.
\newblock
\APACrefbtitle {{SWOT} Data Challenge {NATL60} [Dataset].} {{SWOT} data
  challenge {NATL60} [dataset].}
\newblock
\APACaddressPublisher{}{CNES}.
\newblock
\begin{APACrefDOI} \doi{10.24400/527896/A01-2020.002} \end{APACrefDOI}
\PrintBackRefs{\CurrentBib}

\bibitem [\protect \citeauthoryear {%
CLS/MEOM%
}{%
CLS/MEOM%
}{%
{\protect \APACyear {2021}}%
}]{%
DATAch2021}
\APACinsertmetastar {%
DATAch2021}%
\begin{APACrefauthors}%
CLS/MEOM.%
\end{APACrefauthors}%
\unskip\
\newblock
\APACrefYearMonthDay{2021}{}{}.
\newblock
\APACrefbtitle {DATA CHALLENGE {OSE} - {2021A\_SSH\_MAPPING\_OSE} [Dataset].}
  {Data challenge {OSE} - {2021A\_SSH\_MAPPING\_OSE} [dataset].}
\newblock
\APACaddressPublisher{}{CNES}.
\newblock
\begin{APACrefDOI} \doi{10.24400/527896/a01-2021.005} \end{APACrefDOI}
\PrintBackRefs{\CurrentBib}

\bibitem [\protect \citeauthoryear {%
CMEMS%
}{%
CMEMS%
}{%
{\protect \APACyear {2020}}%
}]{%
DATAglorys}
\APACinsertmetastar {%
DATAglorys}%
\begin{APACrefauthors}%
CMEMS.%
\end{APACrefauthors}%
\unskip\
\newblock
\APACrefYearMonthDay{2020}{}{}.
\newblock
\APACrefbtitle {Global Ocean Physics Reanalysis [Dataset].} {Global ocean
  physics reanalysis [dataset].}
\newblock
\APACaddressPublisher{}{Mercator Ocean International}.
\newblock
\begin{APACrefDOI} \doi{10.48670/moi-00021} \end{APACrefDOI}
\PrintBackRefs{\CurrentBib}

\bibitem [\protect \citeauthoryear {%
CMEMS%
}{%
CMEMS%
}{%
{\protect \APACyear {2021}}%
}]{%
DATAL3tracks}
\APACinsertmetastar {%
DATAL3tracks}%
\begin{APACrefauthors}%
CMEMS.%
\end{APACrefauthors}%
\unskip\
\newblock
\APACrefYearMonthDay{2021}{}{}.
\newblock
\APACrefbtitle {Global Ocean Along-Track {L3} Sea Surface Heights Reprocessed
  (1993-Ongoing) Tailored For Data Assimilation [Dataset].} {Global ocean
  along-track {L3} sea surface heights reprocessed (1993-ongoing) tailored for
  data assimilation [dataset].}
\newblock
\APACaddressPublisher{}{Mercator Ocean International}.
\newblock
\begin{APACrefDOI} \doi{10.48670/MOI-00146} \end{APACrefDOI}
\PrintBackRefs{\CurrentBib}

\bibitem [\protect \citeauthoryear {%
CMEMS%
}{%
CMEMS%
}{%
{\protect \APACyear {2023}}%
}]{%
DATAcloud}
\APACinsertmetastar {%
DATAcloud}%
\begin{APACrefauthors}%
CMEMS.%
\end{APACrefauthors}%
\unskip\
\newblock
\APACrefYearMonthDay{2023}{}{}.
\newblock
\APACrefbtitle {Global Oceans Sea Surface Temperature Multi-sensor {L3}
  Observations [Dataset].} {Global oceans sea surface temperature multi-sensor
  {L3} observations [dataset].}
\newblock
\APACaddressPublisher{}{Mercator Ocean International}.
\newblock
\begin{APACrefDOI} \doi{10.48670/MOI-00164} \end{APACrefDOI}
\PrintBackRefs{\CurrentBib}

\bibitem [\protect \citeauthoryear {%
Donlon%
\ \protect \BOthers {.}}{%
Donlon%
\ \protect \BOthers {.}}{%
{\protect \APACyear {2012}}%
}]{%
donlon2012}
\APACinsertmetastar {%
donlon2012}%
\begin{APACrefauthors}%
Donlon, C\BPBI J.%
, Martin, M.%
, Stark, J.%
, Roberts-Jones, J.%
, Fiedler, E.%
\BCBL {}\ \BBA {} Wimmer, W.%
\end{APACrefauthors}%
\unskip\
\newblock
\APACrefYearMonthDay{2012}{}{}.
\newblock
{\BBOQ}\APACrefatitle {The Operational Sea Surface Temperature and Sea Ice
  Analysis ({OSTIA}) system} {The operational sea surface temperature and sea
  ice analysis ({OSTIA}) system}.{\BBCQ}
\newblock
\APACjournalVolNumPages{Advanced Along Track Scanning Radiometer (AATSR),
  Special Issue of Remote Sensing of Environment}{116}{}{140-158}.
\newblock
\begin{APACrefDOI} \doi{10.1016/j.rse.2010.10.017} \end{APACrefDOI}
\PrintBackRefs{\CurrentBib}

\bibitem [\protect \citeauthoryear {%
Emery%
, Brown%
\BCBL {}\ \BBA {} Nowak%
}{%
Emery%
\ \protect \BOthers {.}}{%
{\protect \APACyear {1989}}%
}]{%
emery1989avhrr}
\APACinsertmetastar {%
emery1989avhrr}%
\begin{APACrefauthors}%
Emery, W\BPBI J.%
, Brown, J.%
\BCBL {}\ \BBA {} Nowak, Z\BPBI P.%
\end{APACrefauthors}%
\unskip\
\newblock
\APACrefYearMonthDay{1989}{}{}.
\newblock
{\BBOQ}\APACrefatitle {{AVHRR} image navigation-Summary and review} {{AVHRR}
  image navigation-summary and review}.{\BBCQ}
\newblock
\APACjournalVolNumPages{Photogrammetric engineering and remote
  sensing}{4}{}{1175--1183}.
\PrintBackRefs{\CurrentBib}

\bibitem [\protect \citeauthoryear {%
Fablet%
, Amar%
, Febvre%
, Beauchamp%
\BCBL {}\ \BBA {} Chapron%
}{%
Fablet%
\ \protect \BOthers {.}}{%
{\protect \APACyear {2021}}%
}]{%
fablet2021}
\APACinsertmetastar {%
fablet2021}%
\begin{APACrefauthors}%
Fablet, R.%
, Amar, M.%
, Febvre, Q.%
, Beauchamp, M.%
\BCBL {}\ \BBA {} Chapron, B.%
\end{APACrefauthors}%
\unskip\
\newblock
\APACrefYearMonthDay{2021}{{\APACmonth{06}}}{}.
\newblock
{\BBOQ}\APACrefatitle {End-to-end physics-informed representation learning for
  satellite ocean remote sensing data: Applications to satellite altimetry and
  sea surface currents} {End-to-end physics-informed representation learning
  for satellite ocean remote sensing data: Applications to satellite altimetry
  and sea surface currents}.{\BBCQ}
\newblock
\APACjournalVolNumPages{ISPRS Annals of the Photogrammetry, Remote Sensing and
  Spatial Information Sciences}{5}{}{295-302}.
\newblock
\begin{APACrefDOI} \doi{10.5194/ISPRS-ANNALS-V-3-2021-295-2021}
  \end{APACrefDOI}
\PrintBackRefs{\CurrentBib}

\bibitem [\protect \citeauthoryear {%
Fablet%
, Febvre%
\BCBL {}\ \BBA {} Chapron%
}{%
Fablet%
\ \protect \BOthers {.}}{%
{\protect \APACyear {2023}}%
}]{%
fablet2023}
\APACinsertmetastar {%
fablet2023}%
\begin{APACrefauthors}%
Fablet, R.%
, Febvre, Q.%
\BCBL {}\ \BBA {} Chapron, B.%
\end{APACrefauthors}%
\unskip\
\newblock
\APACrefYearMonthDay{2023}{}{}.
\newblock
{\BBOQ}\APACrefatitle {Multimodal {4DVarNets} for the Reconstruction of Sea
  Surface Dynamics From {SST-SSH} Synergies} {Multimodal {4DVarNets} for the
  reconstruction of sea surface dynamics from {SST-SSH} synergies}.{\BBCQ}
\newblock
\APACjournalVolNumPages{IEEE Transactions on Geoscience and Remote
  Sensing}{61}{}{}.
\newblock
\begin{APACrefDOI} \doi{10.1109/TGRS.2023.3268006} \end{APACrefDOI}
\PrintBackRefs{\CurrentBib}

\bibitem [\protect \citeauthoryear {%
Filoche%
, Archambault%
, Charantonis%
\BCBL {}\ \BBA {} Béréziat%
}{%
Filoche%
\ \protect \BOthers {.}}{%
{\protect \APACyear {2022}}%
}]{%
filoche2022}
\APACinsertmetastar {%
filoche2022}%
\begin{APACrefauthors}%
Filoche, A.%
, Archambault, T.%
, Charantonis, A.%
\BCBL {}\ \BBA {} Béréziat, D.%
\end{APACrefauthors}%
\unskip\
\newblock
\APACrefYearMonthDay{2022}{}{}.
\newblock
{\BBOQ}\APACrefatitle {Statistics-free interpolation of ocean observations with
  deep spatio-temporal prior} {Statistics-free interpolation of ocean
  observations with deep spatio-temporal prior}.{\BBCQ}
\newblock
\BIn{} \APACrefbtitle {{ECML/PKDD Workshop on Machine Learning for Earth
  Observation and Prediction (MACLEAN)}.} {{ECML/PKDD Workshop on Machine
  Learning for Earth Observation and Prediction (MACLEAN)}.}
\PrintBackRefs{\CurrentBib}

\bibitem [\protect \citeauthoryear {%
Gaultier%
, Ubelmann%
\BCBL {}\ \BBA {} Fu%
}{%
Gaultier%
\ \protect \BOthers {.}}{%
{\protect \APACyear {2016}}%
}]{%
gaultier2016}
\APACinsertmetastar {%
gaultier2016}%
\begin{APACrefauthors}%
Gaultier, L.%
, Ubelmann, C.%
\BCBL {}\ \BBA {} Fu, L.%
\end{APACrefauthors}%
\unskip\
\newblock
\APACrefYearMonthDay{2016}{}{}.
\newblock
{\BBOQ}\APACrefatitle {The challenge of using future {SWOT} data for oceanic
  field reconstruction} {The challenge of using future {SWOT} data for oceanic
  field reconstruction}.{\BBCQ}
\newblock
\APACjournalVolNumPages{Journal of Atmospheric and Oceanic
  Technology}{33}{}{119-126}.
\newblock
\begin{APACrefDOI} \doi{10.1175/JTECH-D-15-0160.1} \end{APACrefDOI}
\PrintBackRefs{\CurrentBib}

\bibitem [\protect \citeauthoryear {%
González-Haro%
, Isern-Fontanet%
, Tandeo%
\BCBL {}\ \BBA {} Garello%
}{%
González-Haro%
\ \protect \BOthers {.}}{%
{\protect \APACyear {2020}}%
}]{%
gonzalez2020}
\APACinsertmetastar {%
gonzalez2020}%
\begin{APACrefauthors}%
González-Haro, C.%
, Isern-Fontanet, J.%
, Tandeo, P.%
\BCBL {}\ \BBA {} Garello, R.%
\end{APACrefauthors}%
\unskip\
\newblock
\APACrefYearMonthDay{2020}{}{}.
\newblock
{\BBOQ}\APACrefatitle {Ocean Surface Currents Reconstruction: Spectral
  Characterization of the Transfer Function Between {SST} and {SSH}} {Ocean
  surface currents reconstruction: Spectral characterization of the transfer
  function between {SST} and {SSH}}.{\BBCQ}
\newblock
\APACjournalVolNumPages{Journal of Geophysical Research:
  Oceans}{125}{10}{e2019JC015958}.
\newblock
\APACrefnote{e2019JC015958 10.1029/2019JC015958}
\newblock
\begin{APACrefDOI} \doi{https://doi.org/10.1029/2019JC015958} \end{APACrefDOI}
\PrintBackRefs{\CurrentBib}

\bibitem [\protect \citeauthoryear {%
Guo%
\ \protect \BOthers {.}}{%
Guo%
\ \protect \BOthers {.}}{%
{\protect \APACyear {2021}}%
}]{%
guo2021}
\APACinsertmetastar {%
guo2021}%
\begin{APACrefauthors}%
Guo, M\BHBI H.%
, Xu, T\BHBI X.%
, Liu, J\BHBI J.%
, Liu, Z\BHBI N.%
, Jiang, P\BHBI T.%
, Mu, T\BHBI J.%
\BDBL {}Hu, S\BHBI M.%
\end{APACrefauthors}%
\unskip\
\newblock
\APACrefYearMonthDay{2021}{{\APACmonth{11}}}{}.
\newblock
{\BBOQ}\APACrefatitle {Attention Mechanisms in Computer Vision: A Survey}
  {Attention mechanisms in computer vision: A survey}.{\BBCQ}
\newblock
\APACjournalVolNumPages{Computational Visual Media}{8}{}{331-368}.
\newblock
\begin{APACrefDOI} \doi{10.1007/s41095-022-0271-y} \end{APACrefDOI}
\PrintBackRefs{\CurrentBib}

\bibitem [\protect \citeauthoryear {%
He%
, Zhang%
, Ren%
\BCBL {}\ \BBA {} Sun%
}{%
He%
\ \protect \BOthers {.}}{%
{\protect \APACyear {2016}}%
}]{%
he2016}
\APACinsertmetastar {%
he2016}%
\begin{APACrefauthors}%
He, K.%
, Zhang, X.%
, Ren, S.%
\BCBL {}\ \BBA {} Sun, J.%
\end{APACrefauthors}%
\unskip\
\newblock
\APACrefYearMonthDay{2016}{}{}.
\newblock
{\BBOQ}\APACrefatitle {Deep Residual Learning for Image Recognition} {Deep
  residual learning for image recognition}.{\BBCQ}
\newblock
\BIn{} \APACrefbtitle {Conference on Computer Vision and Pattern Recognition
  ({CVPR})} {Conference on computer vision and pattern recognition ({CVPR})}\
  (\BPGS\ 770--778).
\newblock
\begin{APACrefDOI} \doi{10.1109/CVPR.2016.90} \end{APACrefDOI}
\PrintBackRefs{\CurrentBib}

\bibitem [\protect \citeauthoryear {%
Hinton%
\ \BBA {} Dean%
}{%
Hinton%
\ \BBA {} Dean%
}{%
{\protect \APACyear {2015}}%
}]{%
Hinton2015}
\APACinsertmetastar {%
Hinton2015}%
\begin{APACrefauthors}%
Hinton, G.%
\BCBT {}\ \BBA {} Dean, J.%
\end{APACrefauthors}%
\unskip\
\newblock
\APACrefYearMonthDay{2015}{}{}.
\newblock
{\BBOQ}\APACrefatitle {Distilling the Knowledge in a Neural Network}
  {Distilling the knowledge in a neural network}.{\BBCQ}
\newblock
\BIn{} \APACrefbtitle {{NIPS Deep Learning and Representation Learning
  Workshop}.} {{NIPS Deep Learning and Representation Learning Workshop}.}
\PrintBackRefs{\CurrentBib}

\bibitem [\protect \citeauthoryear {%
Isern-Fontanet%
, Chapron%
, Lapeyre%
\BCBL {}\ \BBA {} Klein%
}{%
Isern-Fontanet%
\ \protect \BOthers {.}}{%
{\protect \APACyear {2006}}%
}]{%
isernfontanet2006}
\APACinsertmetastar {%
isernfontanet2006}%
\begin{APACrefauthors}%
Isern-Fontanet, J.%
, Chapron, B.%
, Lapeyre, G.%
\BCBL {}\ \BBA {} Klein, P.%
\end{APACrefauthors}%
\unskip\
\newblock
\APACrefYearMonthDay{2006}{}{}.
\newblock
{\BBOQ}\APACrefatitle {Potential use of microwave sea surface temperatures for
  the estimation of ocean currents} {Potential use of microwave sea surface
  temperatures for the estimation of ocean currents}.{\BBCQ}
\newblock
\APACjournalVolNumPages{Geophys. Res. Lett}{33}{}{24608}.
\newblock
\begin{APACrefDOI} \doi{10.1029/2006GL027801} \end{APACrefDOI}
\PrintBackRefs{\CurrentBib}

\bibitem [\protect \citeauthoryear {%
Jam%
\ \protect \BOthers {.}}{%
Jam%
\ \protect \BOthers {.}}{%
{\protect \APACyear {2021}}%
}]{%
jam2021}
\APACinsertmetastar {%
jam2021}%
\begin{APACrefauthors}%
Jam, J.%
, Kendrick, C.%
, Walker, K.%
, Drouard, V.%
, Hsu, J.%
\BCBL {}\ \BBA {} Yap, M.%
\end{APACrefauthors}%
\unskip\
\newblock
\APACrefYearMonthDay{2021}{{\APACmonth{02}}}{}.
\newblock
{\BBOQ}\APACrefatitle {A comprehensive review of past and present image
  inpainting methods} {A comprehensive review of past and present image
  inpainting methods}.{\BBCQ}
\newblock
\APACjournalVolNumPages{Computer Vision and Image Understanding}{203}{}{}.
\newblock
\begin{APACrefDOI} \doi{10.1016/J.CVIU.2020.103147} \end{APACrefDOI}
\PrintBackRefs{\CurrentBib}

\bibitem [\protect \citeauthoryear {%
Jayne%
\ \BBA {} Marotzke%
}{%
Jayne%
\ \BBA {} Marotzke%
}{%
{\protect \APACyear {2002}}%
}]{%
jayne2002}
\APACinsertmetastar {%
jayne2002}%
\begin{APACrefauthors}%
Jayne, S.%
\BCBT {}\ \BBA {} Marotzke, J.%
\end{APACrefauthors}%
\unskip\
\newblock
\APACrefYearMonthDay{2002}{{\APACmonth{12}}}{}.
\newblock
{\BBOQ}\APACrefatitle {The Oceanic Eddy Heat Transport} {The oceanic eddy heat
  transport}.{\BBCQ}
\newblock
\APACjournalVolNumPages{Journal of Physical Oceanography}{32}{}{3328-3345}.
\newblock
\begin{APACrefDOI} \doi{10.1175/1520-0485(2002)032<3328:TOEHT>2.0.CO;2}
  \end{APACrefDOI}
\PrintBackRefs{\CurrentBib}

\bibitem [\protect \citeauthoryear {%
Kahru%
, Di~Lorenzo%
, Manzano-Sarabia%
\BCBL {}\ \BBA {} Mitchell%
}{%
Kahru%
\ \protect \BOthers {.}}{%
{\protect \APACyear {2012}}%
}]{%
Kahru2012}
\APACinsertmetastar {%
Kahru2012}%
\begin{APACrefauthors}%
Kahru, M.%
, Di~Lorenzo, E.%
, Manzano-Sarabia, M.%
\BCBL {}\ \BBA {} Mitchell, B\BPBI G.%
\end{APACrefauthors}%
\unskip\
\newblock
\APACrefYearMonthDay{2012}{{\APACmonth{03}}}{}.
\newblock
{\BBOQ}\APACrefatitle {{Spatial and temporal statistics of sea surface
  temperature and chlorophyll fronts in the California Current}} {{Spatial and
  temporal statistics of sea surface temperature and chlorophyll fronts in the
  California Current}}.{\BBCQ}
\newblock
\APACjournalVolNumPages{Journal of Plankton Research}{34}{9}{749-760}.
\newblock
\begin{APACrefDOI} \doi{10.1093/plankt/fbs010} \end{APACrefDOI}
\PrintBackRefs{\CurrentBib}

\bibitem [\protect \citeauthoryear {%
Kang%
, Curchitser%
\BCBL {}\ \BBA {} Rosati%
}{%
Kang%
\ \protect \BOthers {.}}{%
{\protect \APACyear {2016}}%
}]{%
kang2016}
\APACinsertmetastar {%
kang2016}%
\begin{APACrefauthors}%
Kang, D.%
, Curchitser, E\BPBI N.%
\BCBL {}\ \BBA {} Rosati, A.%
\end{APACrefauthors}%
\unskip\
\newblock
\APACrefYearMonthDay{2016}{}{}.
\newblock
{\BBOQ}\APACrefatitle {Seasonal Variability of the {Gulf Stream} Kinetic
  Energy} {Seasonal variability of the {Gulf Stream} kinetic energy}.{\BBCQ}
\newblock
\APACjournalVolNumPages{Journal of Physical Oceanography}{46}{4}{1189 - 1207}.
\newblock
\begin{APACrefDOI} \doi{https://doi.org/10.1175/JPO-D-15-0235.1}
  \end{APACrefDOI}
\PrintBackRefs{\CurrentBib}

\bibitem [\protect \citeauthoryear {%
Kingma%
\ \BBA {} Ba%
}{%
Kingma%
\ \BBA {} Ba%
}{%
{\protect \APACyear {2017}}%
}]{%
kingma2017adam}
\APACinsertmetastar {%
kingma2017adam}%
\begin{APACrefauthors}%
Kingma, D\BPBI P.%
\BCBT {}\ \BBA {} Ba, J.%
\end{APACrefauthors}%
\unskip\
\newblock
\APACrefYearMonthDay{2017}{}{}.
\newblock
\APACrefbtitle {Adam: A Method for Stochastic Optimization.} {Adam: A method
  for stochastic optimization.}
\PrintBackRefs{\CurrentBib}

\bibitem [\protect \citeauthoryear {%
Le~Guillou%
\ \protect \BOthers {.}}{%
Le~Guillou%
\ \protect \BOthers {.}}{%
{\protect \APACyear {2020}}%
}]{%
guillou2020_bfn}
\APACinsertmetastar {%
guillou2020_bfn}%
\begin{APACrefauthors}%
Le~Guillou, F.%
, Metref, S.%
, Cosme, E.%
, Ubelmann, C.%
, Ballarotta, M.%
, Verron, J.%
\BCBL {}\ \BBA {} Le~Sommer, J.%
\end{APACrefauthors}%
\unskip\
\newblock
\APACrefYearMonthDay{2020}{{\APACmonth{10}}}{}.
\newblock
\APACrefbtitle {Mapping altimetry in the forthcoming {SWOT} era by
  back-and-forth nudging a one-layer quasi-geostrophic model.} {Mapping
  altimetry in the forthcoming {SWOT} era by back-and-forth nudging a one-layer
  quasi-geostrophic model.}
\newblock
\APAChowpublished {Earth and Space Science Open Archive}.
\newblock
\begin{APACrefDOI} \doi{10.1002/ESSOAR.10504575.1} \end{APACrefDOI}
\PrintBackRefs{\CurrentBib}

\bibitem [\protect \citeauthoryear {%
Madec%
\ \protect \BOthers {.}}{%
Madec%
\ \protect \BOthers {.}}{%
{\protect \APACyear {2017}}%
}]{%
madec2017nemo}
\APACinsertmetastar {%
madec2017nemo}%
\begin{APACrefauthors}%
Madec, G.%
, Bourdall{\'e}-Badie, R.%
, Bouttier, P\BHBI A.%
, Bricaud, C.%
, Bruciaferri, D.%
, Calvert, D.%
\BDBL {}others%
\end{APACrefauthors}%
\unskip\
\newblock
\APACrefYearMonthDay{2017}{}{}.
\newblock
\APACrefbtitle {NEMO ocean engine.} {Nemo ocean engine.}
\PrintBackRefs{\CurrentBib}

\bibitem [\protect \citeauthoryear {%
Manucharyan%
, Siegelman%
\BCBL {}\ \BBA {} Klein%
}{%
Manucharyan%
\ \protect \BOthers {.}}{%
{\protect \APACyear {2020}}%
}]{%
manucharyan2020}
\APACinsertmetastar {%
manucharyan2020}%
\begin{APACrefauthors}%
Manucharyan, G.%
, Siegelman, L.%
\BCBL {}\ \BBA {} Klein, P.%
\end{APACrefauthors}%
\unskip\
\newblock
\APACrefYearMonthDay{2020}{{\APACmonth{01}}}{}.
\newblock
{\BBOQ}\APACrefatitle {A Deep Learning Approach to Spatiotemporal Sea Surface
  Height Interpolation and Estimation of Deep Currents in Geostrophic Ocean
  Turbulence} {A deep learning approach to spatiotemporal sea surface height
  interpolation and estimation of deep currents in geostrophic ocean
  turbulence}.{\BBCQ}
\newblock
\APACjournalVolNumPages{Journal of Advances in Modeling Earth Systems}{13}{}{}.
\newblock
\begin{APACrefDOI} \doi{10.1029/2019MS001965} \end{APACrefDOI}
\PrintBackRefs{\CurrentBib}

\bibitem [\protect \citeauthoryear {%
S.~Martin%
}{%
S.~Martin%
}{%
{\protect \APACyear {2014}}%
}]{%
martin_2014}
\APACinsertmetastar {%
martin_2014}%
\begin{APACrefauthors}%
Martin, S.%
\end{APACrefauthors}%
\unskip\
\newblock
\APACrefYear{2014}.
\newblock
\APACrefbtitle {An Introduction to Ocean Remote Sensing} {An introduction to
  ocean remote sensing}\ (\PrintOrdinal{2}\ \BEd).
\newblock
\APACaddressPublisher{}{Cambridge University Press}.
\newblock
\begin{APACrefDOI} \doi{10.1017/CBO9781139094368} \end{APACrefDOI}
\PrintBackRefs{\CurrentBib}

\bibitem [\protect \citeauthoryear {%
S\BPBI A.~Martin%
, Manucharyan%
\BCBL {}\ \BBA {} Klein%
}{%
S\BPBI A.~Martin%
\ \protect \BOthers {.}}{%
{\protect \APACyear {2023}}%
}]{%
martin2023}
\APACinsertmetastar {%
martin2023}%
\begin{APACrefauthors}%
Martin, S\BPBI A.%
, Manucharyan, G\BPBI E.%
\BCBL {}\ \BBA {} Klein, P.%
\end{APACrefauthors}%
\unskip\
\newblock
\APACrefYearMonthDay{2023}{}{}.
\newblock
{\BBOQ}\APACrefatitle {Synthesizing Sea Surface Temperature and Satellite
  Altimetry Observations Using Deep Learning Improves the Accuracy and
  Resolution of Gridded Sea Surface Height Anomalies} {Synthesizing sea surface
  temperature and satellite altimetry observations using deep learning improves
  the accuracy and resolution of gridded sea surface height anomalies}.{\BBCQ}
\newblock
\APACjournalVolNumPages{Journal of Advances in Modeling Earth
  Systems}{15}{5}{}.
\newblock
\begin{APACrefDOI} \doi{10.1029/2022MS003589} \end{APACrefDOI}
\PrintBackRefs{\CurrentBib}

\bibitem [\protect \citeauthoryear {%
McCann%
, Jin%
\BCBL {}\ \BBA {} Unser%
}{%
McCann%
\ \protect \BOthers {.}}{%
{\protect \APACyear {2017}}%
}]{%
mccann2017}
\APACinsertmetastar {%
mccann2017}%
\begin{APACrefauthors}%
McCann, M.%
, Jin, K.%
\BCBL {}\ \BBA {} Unser, M.%
\end{APACrefauthors}%
\unskip\
\newblock
\APACrefYearMonthDay{2017}{{\APACmonth{11}}}{}.
\newblock
{\BBOQ}\APACrefatitle {Convolutional neural networks for inverse problems in
  imaging: A review} {Convolutional neural networks for inverse problems in
  imaging: A review}.{\BBCQ}
\newblock
\APACjournalVolNumPages{IEEE Signal Processing Magazine}{34}{}{85--95}.
\newblock
\begin{APACrefDOI} \doi{10.1109/MSP.2017.2739299} \end{APACrefDOI}
\PrintBackRefs{\CurrentBib}

\bibitem [\protect \citeauthoryear {%
Mkhinini%
\ \protect \BOthers {.}}{%
Mkhinini%
\ \protect \BOthers {.}}{%
{\protect \APACyear {2014}}%
}]{%
mkhinini2014}
\APACinsertmetastar {%
mkhinini2014}%
\begin{APACrefauthors}%
Mkhinini, N.%
, Coimbra, A\BPBI L\BPBI S.%
, Stegner, A.%
, Arsouze, T.%
, Taupier-Letage, I.%
\BCBL {}\ \BBA {} Béranger, K.%
\end{APACrefauthors}%
\unskip\
\newblock
\APACrefYearMonthDay{2014}{{\APACmonth{12}}}{}.
\newblock
{\BBOQ}\APACrefatitle {Long-lived mesoscale eddies in the eastern
  {Mediterranean Sea}: Analysis of 20 years of {AVISO} geostrophic velocities}
  {Long-lived mesoscale eddies in the eastern {Mediterranean Sea}: Analysis of
  20 years of {AVISO} geostrophic velocities}.{\BBCQ}
\newblock
\APACjournalVolNumPages{Journal of Geophysical Research:
  Oceans}{119}{}{8603--8626}.
\newblock
\begin{APACrefDOI} \doi{10.1002/2014JC010176} \end{APACrefDOI}
\PrintBackRefs{\CurrentBib}

\bibitem [\protect \citeauthoryear {%
Moschos%
, Schwander%
, Stegner%
\BCBL {}\ \BBA {} Gallinari%
}{%
Moschos%
\ \protect \BOthers {.}}{%
{\protect \APACyear {2020}}%
}]{%
moschos2020}
\APACinsertmetastar {%
moschos2020}%
\begin{APACrefauthors}%
Moschos, E.%
, Schwander, O.%
, Stegner, A.%
\BCBL {}\ \BBA {} Gallinari, P.%
\end{APACrefauthors}%
\unskip\
\newblock
\APACrefYearMonthDay{2020}{}{}.
\newblock
{\BBOQ}\APACrefatitle {{Deep-SST-Eddies}: A Deep Learning Framework to Detect
  Oceanic Eddies in Sea Surface Temperature Images} {{Deep-SST-Eddies}: A deep
  learning framework to detect oceanic eddies in sea surface temperature
  images}.{\BBCQ}
\newblock
\BIn{} \APACrefbtitle {{International Conference on Acoustics, Speech and
  Signal Processing (ICASSP)}} {{International Conference on Acoustics, Speech
  and Signal Processing (ICASSP)}}\ (\BPG~4307-4311).
\newblock
\begin{APACrefDOI} \doi{10.1109/ICASSP40776.2020.9053909} \end{APACrefDOI}
\PrintBackRefs{\CurrentBib}

\bibitem [\protect \citeauthoryear {%
Nardelli%
, Cavaliere%
, Charles%
\BCBL {}\ \BBA {} Ciani%
}{%
Nardelli%
\ \protect \BOthers {.}}{%
{\protect \APACyear {2022}}%
}]{%
nardelli2022}
\APACinsertmetastar {%
nardelli2022}%
\begin{APACrefauthors}%
Nardelli, B.%
, Cavaliere, D.%
, Charles, E.%
\BCBL {}\ \BBA {} Ciani, D.%
\end{APACrefauthors}%
\unskip\
\newblock
\APACrefYearMonthDay{2022}{{\APACmonth{02}}}{}.
\newblock
{\BBOQ}\APACrefatitle {Super-Resolving Ocean Dynamics from Space with Computer
  Vision Algorithms} {Super-resolving ocean dynamics from space with computer
  vision algorithms}.{\BBCQ}
\newblock
\APACjournalVolNumPages{Remote Sensing}{14}{}{1159}.
\newblock
\begin{APACrefDOI} \doi{10.3390/RS14051159} \end{APACrefDOI}
\PrintBackRefs{\CurrentBib}

\bibitem [\protect \citeauthoryear {%
NASA/JPL%
}{%
NASA/JPL%
}{%
{\protect \APACyear {2019}}%
}]{%
DATAmur}
\APACinsertmetastar {%
DATAmur}%
\begin{APACrefauthors}%
NASA/JPL.%
\end{APACrefauthors}%
\unskip\
\newblock
\APACrefYearMonthDay{2019}{}{}.
\newblock
\APACrefbtitle {{GHRSST} Level 4 MUR 0.25deg Global Foundation Sea Surface
  Temperature Analysis (v4.2) [Dataset].} {{GHRSST} level 4 mur 0.25deg global
  foundation sea surface temperature analysis (v4.2) [dataset].}
\newblock
\APACaddressPublisher{}{Mercator Ocean International}.
\newblock
\begin{APACrefDOI} \doi{10.5067/GHM25-4FJ42} \end{APACrefDOI}
\PrintBackRefs{\CurrentBib}

\bibitem [\protect \citeauthoryear {%
Ongie%
\ \protect \BOthers {.}}{%
Ongie%
\ \protect \BOthers {.}}{%
{\protect \APACyear {2020}}%
}]{%
ongie2020}
\APACinsertmetastar {%
ongie2020}%
\begin{APACrefauthors}%
Ongie, O.%
, Jalal, A.%
, Metzler, C.%
, Baraniuk, R.%
, Dimakis, A.%
\BCBL {}\ \BBA {} Willett, R.%
\end{APACrefauthors}%
\unskip\
\newblock
\APACrefYearMonthDay{2020}{{\APACmonth{05}}}{}.
\newblock
{\BBOQ}\APACrefatitle {Deep Learning Techniques for Inverse Problems in
  Imaging} {Deep learning techniques for inverse problems in imaging}.{\BBCQ}
\newblock
\APACjournalVolNumPages{IEEE Journal on Selected Areas in Information
  Theory}{1}{}{39--56}.
\PrintBackRefs{\CurrentBib}

\bibitem [\protect \citeauthoryear {%
Pan%
\ \BBA {} Yang%
}{%
Pan%
\ \BBA {} Yang%
}{%
{\protect \APACyear {2010}}%
}]{%
pan2010}
\APACinsertmetastar {%
pan2010}%
\begin{APACrefauthors}%
Pan, S\BPBI J.%
\BCBT {}\ \BBA {} Yang, Q.%
\end{APACrefauthors}%
\unskip\
\newblock
\APACrefYearMonthDay{2010}{}{}.
\newblock
{\BBOQ}\APACrefatitle {A Survey on Transfer Learning} {A survey on transfer
  learning}.{\BBCQ}
\newblock
\APACjournalVolNumPages{IEEE Transactions on Knowledge and Data
  Engineering}{22}{10}{1345--1359}.
\newblock
\begin{APACrefDOI} \doi{10.1109/TKDE.2009.191} \end{APACrefDOI}
\PrintBackRefs{\CurrentBib}

\bibitem [\protect \citeauthoryear {%
Qin%
, Zeng%
, Zong%
\BCBL {}\ \BBA {} Xu%
}{%
Qin%
\ \protect \BOthers {.}}{%
{\protect \APACyear {2021}}%
}]{%
qin2021}
\APACinsertmetastar {%
qin2021}%
\begin{APACrefauthors}%
Qin, Z.%
, Zeng, Q.%
, Zong, Y.%
\BCBL {}\ \BBA {} Xu, F.%
\end{APACrefauthors}%
\unskip\
\newblock
\APACrefYearMonthDay{2021}{{\APACmonth{09}}}{}.
\newblock
{\BBOQ}\APACrefatitle {Image inpainting based on deep learning: A review}
  {Image inpainting based on deep learning: A review}.{\BBCQ}
\newblock
\APACjournalVolNumPages{Displays}{69}{}{102028}.
\newblock
\begin{APACrefDOI} \doi{10.1016/J.DISPLA.2021.102028} \end{APACrefDOI}
\PrintBackRefs{\CurrentBib}

\bibitem [\protect \citeauthoryear {%
Stegner%
\ \protect \BOthers {.}}{%
Stegner%
\ \protect \BOthers {.}}{%
{\protect \APACyear {2021}}%
}]{%
stegner2021}
\APACinsertmetastar {%
stegner2021}%
\begin{APACrefauthors}%
Stegner, A.%
, Le~Vu, B.%
, Dumas, F.%
, Ghannami, M.%
, Nicolle, A.%
, Durand, C.%
\BCBL {}\ \BBA {} Faugere, Y.%
\end{APACrefauthors}%
\unskip\
\newblock
\APACrefYearMonthDay{2021}{{\APACmonth{09}}}{}.
\newblock
{\BBOQ}\APACrefatitle {Cyclone-Anticyclone Asymmetry of Eddy Detection on
  Gridded Altimetry Product in the Mediterranean Sea} {Cyclone-anticyclone
  asymmetry of eddy detection on gridded altimetry product in the mediterranean
  sea}.{\BBCQ}
\newblock
\APACjournalVolNumPages{Journal of Geophysical Research: Oceans}{126}{}{}.
\newblock
\begin{APACrefDOI} \doi{10.1029/2021JC017475} \end{APACrefDOI}
\PrintBackRefs{\CurrentBib}

\bibitem [\protect \citeauthoryear {%
Taburet%
\ \protect \BOthers {.}}{%
Taburet%
\ \protect \BOthers {.}}{%
{\protect \APACyear {2019}}%
}]{%
taburet2019}
\APACinsertmetastar {%
taburet2019}%
\begin{APACrefauthors}%
Taburet, G.%
, Sanchez-Roman, A.%
, Ballarotta, M.%
, Pujol, M\BHBI I.%
, Legeais, J\BHBI F.%
, Fournier, F.%
\BDBL {}Dibarboure, G.%
\end{APACrefauthors}%
\unskip\
\newblock
\APACrefYearMonthDay{2019}{}{}.
\newblock
{\BBOQ}\APACrefatitle {{DUACS DT2018}: 25 years of reprocessed sea level
  altimetry products} {{DUACS DT2018}: 25 years of reprocessed sea level
  altimetry products}.{\BBCQ}
\newblock
\APACjournalVolNumPages{Ocean Sci}{15}{}{1207-1224}.
\newblock
\begin{APACrefURL} \url{https://doi.org/10.5194/os-15-1207-2019}
  \end{APACrefURL}
\newblock
\begin{APACrefDOI} \doi{10.5194/os-15-1207-2019} \end{APACrefDOI}
\PrintBackRefs{\CurrentBib}

\bibitem [\protect \citeauthoryear {%
Thiria%
\ \protect \BOthers {.}}{%
Thiria%
\ \protect \BOthers {.}}{%
{\protect \APACyear {2023}}%
}]{%
resac}
\APACinsertmetastar {%
resac}%
\begin{APACrefauthors}%
Thiria, S.%
, Sorror, C.%
, Archambault, T.%
, Charantonis, A.%
, Béréziat, D.%
, Mejia, C.%
\BDBL {}Crepon, M.%
\end{APACrefauthors}%
\unskip\
\newblock
\APACrefYearMonthDay{2023}{}{}.
\newblock
{\BBOQ}\APACrefatitle {Downscaling of ocean fields by fusion of heterogeneous
  observations using Deep Learning algorithms} {Downscaling of ocean fields by
  fusion of heterogeneous observations using deep learning algorithms}.{\BBCQ}
\newblock
\APACjournalVolNumPages{Ocean Modeling}{182}{}{}.
\PrintBackRefs{\CurrentBib}

\bibitem [\protect \citeauthoryear {%
Ubelmann%
, Cornuelle%
\BCBL {}\ \BBA {} Fu%
}{%
Ubelmann%
\ \protect \BOthers {.}}{%
{\protect \APACyear {2016}}%
}]{%
ubelmann2016_dymost}
\APACinsertmetastar {%
ubelmann2016_dymost}%
\begin{APACrefauthors}%
Ubelmann, C.%
, Cornuelle, B.%
\BCBL {}\ \BBA {} Fu, L.%
\end{APACrefauthors}%
\unskip\
\newblock
\APACrefYearMonthDay{2016}{{\APACmonth{08}}}{}.
\newblock
{\BBOQ}\APACrefatitle {Dynamic Mapping of Along-Track Ocean Altimetry: Method
  and Performance from Observing System Simulation Experiments} {Dynamic
  mapping of along-track ocean altimetry: Method and performance from observing
  system simulation experiments}.{\BBCQ}
\newblock
\APACjournalVolNumPages{Journal of Atmospheric and Oceanic
  Technology}{33}{}{1691-1699}.
\newblock
\begin{APACrefDOI} \doi{10.1175/JTECH-D-15-0163.1} \end{APACrefDOI}
\PrintBackRefs{\CurrentBib}

\bibitem [\protect \citeauthoryear {%
Vu%
, Stegner%
\BCBL {}\ \BBA {} Arsouze%
}{%
Vu%
\ \protect \BOthers {.}}{%
{\protect \APACyear {2018}}%
}]{%
AMEDA}
\APACinsertmetastar {%
AMEDA}%
\begin{APACrefauthors}%
Vu, B\BPBI L.%
, Stegner, A.%
\BCBL {}\ \BBA {} Arsouze, T.%
\end{APACrefauthors}%
\unskip\
\newblock
\APACrefYearMonthDay{2018}{{\APACmonth{04}}}{}.
\newblock
{\BBOQ}\APACrefatitle {Angular Momentum Eddy Detection and Tracking Algorithm
  ({AMEDA}) and Its Application to Coastal Eddy Formation} {Angular momentum
  eddy detection and tracking algorithm ({AMEDA}) and its application to
  coastal eddy formation}.{\BBCQ}
\newblock
\APACjournalVolNumPages{Journal of Atmospheric and Oceanic
  Technology}{35}{}{739-762}.
\newblock
\begin{APACrefDOI} \doi{10.1175/JTECH-D-17-0010.1} \end{APACrefDOI}
\PrintBackRefs{\CurrentBib}

\bibitem [\protect \citeauthoryear {%
Woo%
, Park%
, Lee%
\BCBL {}\ \BBA {} Kweon%
}{%
Woo%
\ \protect \BOthers {.}}{%
{\protect \APACyear {2018}}%
}]{%
woo2018}
\APACinsertmetastar {%
woo2018}%
\begin{APACrefauthors}%
Woo, S.%
, Park, J.%
, Lee, J\BHBI Y.%
\BCBL {}\ \BBA {} Kweon, I\BPBI S.%
\end{APACrefauthors}%
\unskip\
\newblock
\APACrefYearMonthDay{2018}{}{}.
\newblock
{\BBOQ}\APACrefatitle {{CBAM}: Convolutional Block Attention Module} {{CBAM}:
  Convolutional block attention module}.{\BBCQ}
\newblock
\APACjournalVolNumPages{Computer Vision and Pattern Recognition}{}{}{}.
\PrintBackRefs{\CurrentBib}

\bibitem [\protect \citeauthoryear {%
Zhai%
, Greatbatch%
\BCBL {}\ \BBA {} Kohlmann%
}{%
Zhai%
\ \protect \BOthers {.}}{%
{\protect \APACyear {2008}}%
}]{%
zhai2008}
\APACinsertmetastar {%
zhai2008}%
\begin{APACrefauthors}%
Zhai, X.%
, Greatbatch, R\BPBI J.%
\BCBL {}\ \BBA {} Kohlmann, J\BHBI D.%
\end{APACrefauthors}%
\unskip\
\newblock
\APACrefYearMonthDay{2008}{}{}.
\newblock
{\BBOQ}\APACrefatitle {On the seasonal variability of eddy kinetic energy in
  the {Gulf Stream} region} {On the seasonal variability of eddy kinetic energy
  in the {Gulf Stream} region}.{\BBCQ}
\newblock
\APACjournalVolNumPages{Geophysical Research Letters}{35}{24}{}.
\newblock
\begin{APACrefDOI} \doi{10.1029/2008GL036412} \end{APACrefDOI}
\PrintBackRefs{\CurrentBib}

\end{thebibliography}

\end{document}